\pdfoutput=1  % arXiv: build with pdflatex (figures are PDF/PNG)

\PassOptionsToPackage{numbers,square,sort&compress}{natbib}
\documentclass[Afour,sageh,times,square,sort&compress,numbers]{sagej}
\bibpunct{[}{]}{,}{n}{}{,}   % numeric citations in square brackets
\newif\ifblind
\blindfalse           % camera-ready: authors shown

\newcommand{\repourl}{https://github.com/mohammadi-hadi/xnlp-survey}
\newcommand{\anonrepourl}{https://anonymous.4open.science/r/xnlp-survey-XXXX}
\usepackage{moreverb,url}

\usepackage[colorlinks,bookmarksopen,bookmarksnumbered,citecolor=red,urlcolor=red]{hyperref}

\usepackage{caption}
\usepackage{rotating}
\usepackage{tabularx}
\usepackage{booktabs}
\usepackage{lscape}
\usepackage{longtable}
\usepackage{hyperref}

\usepackage{float} 
\usepackage{graphicx}
\usepackage{adjustbox} 
\usepackage{tabulary} 
\usepackage{natbib}
\usepackage{subcaption}
\usepackage{tikz}
\usetikzlibrary{positioning}
\usepackage{multirow}
\usepackage{enumitem}
\newcolumntype{P}[1]{>{\raggedright\arraybackslash\hspace{0pt}}p{#1}}
\newcommand{\wordcount}{\immediate\write18{texcount -1 -sum -merge \jobname.tex > \jobname.wc }%
  \input{\jobname.wc}}

\newcommand\BibTeX{{\rmfamily B\kern-.05em \textsc{i\kern-.025em b}\kern-.08em
T\kern-.1667em\lower.7ex\hbox{E}\kern-.125emX}}

\def\volumeyear{2025}

\begin{document}

% ======================= FRONT MATTER =========================
\title{Explainability in Practice: A Survey of Explainable NLP Across Various Domains}

\ifblind
  % ------- BLINDED VERSION -------
  \runninghead{BLINDED MANUSCRIPT}
  \author{Anonymous}
  \affiliation{} % leave empty
  \corrauth{}
  \email{}
\else
  % ------- CAMERA-READY VERSION -------
  \runninghead{Mohammadi et al}
  \author{
  Hadi Mohammadi\affilnum{1},
  Robert A. Bagheri\affilnum{1},
  Anastasia Giachanou\affilnum{1},
  Daniel L.~Oberski\affilnum{1}}
  \affiliation{\affilnum{1}Department of Methodology and Statistics, Utrecht University, The Netherlands}
  \corrauth{Hadi Mohammadi, Padualaan 14, 3584 CH Utrecht, The Netherlands.}
  \email{h.mohammadi@uu.nl}
\fi
% ==============================================================

%Words in manuscript: \wordcount

\begin{abstract}
Natural Language Processing (NLP) is now embedded in critical sectors including healthcare, finance, and customer relationship management, where models such as GPT-4o, Gemini, and BERT increasingly inform decisions. The black-box nature of these models has created an urgent need for transparency. This review examines explainable NLP (XNLP) as it is actually deployed, working through seven application domains: medicine, finance, systematic reviews, customer relationship management, chatbots, social and behavioral science, and human resources. For each domain, we ask what kind of explanation the setting needs, which methods are used there, and how they are evaluated. A structured cross-domain synthesis then contrasts how those requirements diverge. We compare the main explanation method families on scope, evidence of faithfulness, and computational cost. We also propose a two-tier evaluation protocol that separates a shared technical core of metrics from the domain-specific validation layer through which those metrics have to be read. The review also addresses areas that remain underrepresented in the XNLP literature, including real-world applicability, the gap between fidelity and faithfulness, and the role of human judgment in assessing explanations. It closes with research directions, among them personalized explanations, human-in-the-loop evaluation, and mechanistic interpretability for large language models.
\end{abstract}


\maketitle

\section{Introduction}
\label{sec:introduction}

Natural Language Processing (NLP) and, more recently, Large Language Models (LLMs) have transformed machine-human interaction by enabling systems to process and generate human language more effectively. Although newer models like OpenAI's GPT-4o and Google's Gemini have pushed the boundaries of language understanding, earlier architectures such as BERT~\cite{devlin2019bert} continue to influence modern NLP pipelines. Indeed, these models have found applications across diverse domains, including healthcare, finance, and Customer Relationship Management (CRM)~\cite{tyagi2023demystifying}, leading to reduced processing times and greater automation. For instance, a study by Oniani et al.~\cite{oniani2024enhancinglargelanguagemodels} showed that using Clinical Practice Guidelines (CPGs) in conjunction with LLMs can produce more precise and contextually relevant treatment recommendations, thus improving clinical decision support (CDS)~\cite{Li2023MedDMLC}. By streamlining such processes, NLP systems offer clear benefits, including rapid analysis, user-friendly interfaces, and the ability to handle substantial amounts of data efficiently~\cite{zhao_survey_2023}.

Despite these advances, most high-performing NLP models operate as ``black boxes.'' The underlying challenge is not merely the large number of parameters in models like GPT-4o or Gemini, but the lack of transparent, human-interpretable decision pathways in neural architectures~\cite{rudin2019we}. In simpler models such as linear regression, it is relatively easier to track how each input feature contributes to a prediction. By contrast, deep neural networks capture complex, non-linear relationships that are difficult to decode, whether they contain millions or billions of parameters. These networks are also frequently trained on massive datasets where historical or societal biases may be embedded~\cite{bolukbasi2016man, caliskan2017semantics}, creating a risk of perpetuating discrimination in downstream predictions~\cite{barocas2016big}. In practice, such biases have been detected in various sectors, including hiring algorithms~\cite{raghavan2020mitigating}, medical diagnostics~\cite{igoe2021algorithmic}, and financial services~\cite{kozodoi2022fairness}. When left unexamined, these biases can lead to adverse outcomes, such as unfair treatment of job applicants, unequal access to credit, or suboptimal healthcare recommendations.

\begin{figure}[!t]
  \centering
  \includegraphics[width=0.99\linewidth]{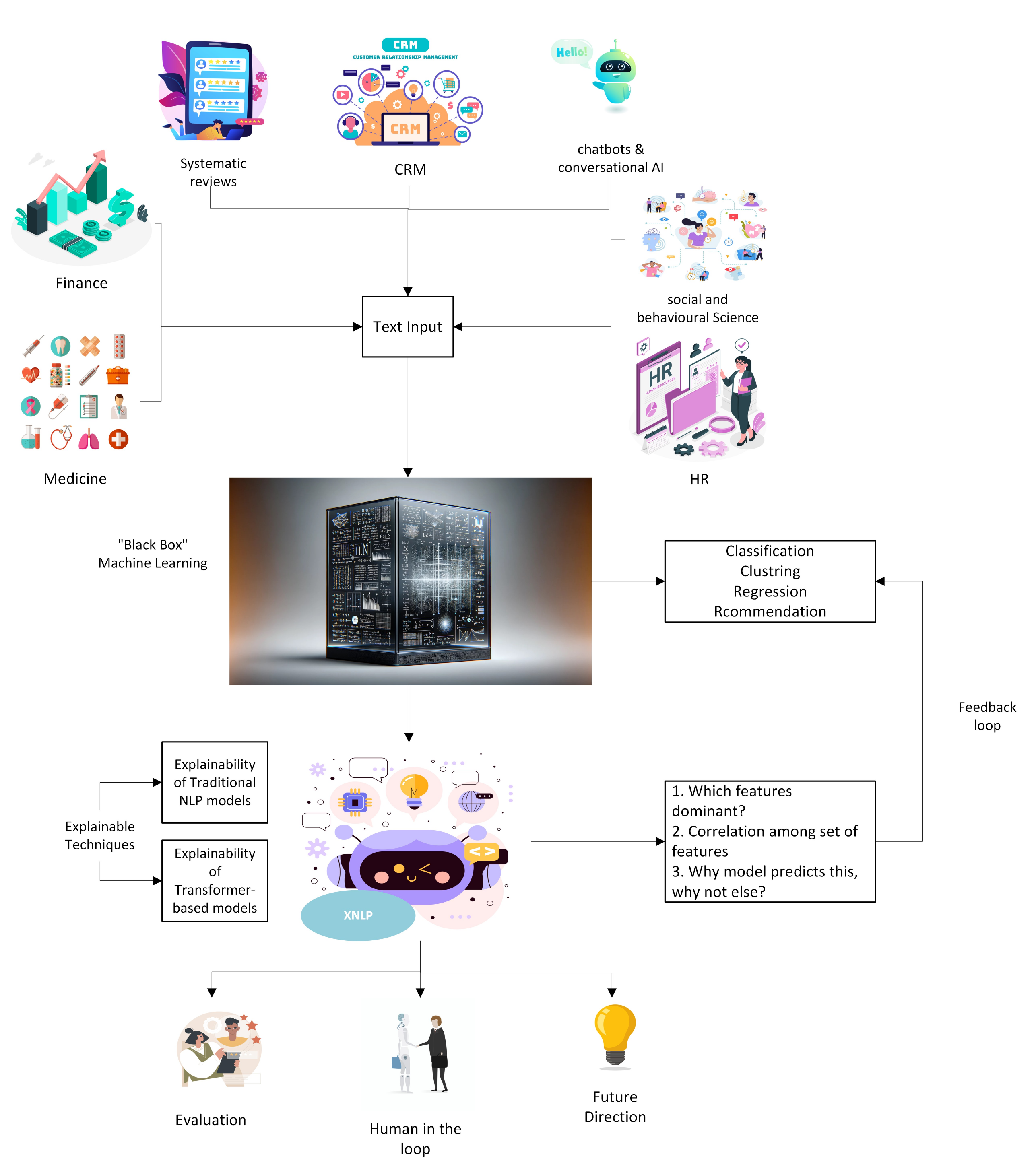}
  \caption{Pipeline for XNLP. A model processes the input text, and an explanation layer highlights the tokens or internal representations behind the output.}
  \label{fig:xnlp_pipeline}
\end{figure}

Explainable AI (XAI) initiatives provide a collection of methods and tools to make these ``black-box'' processes more transparent~\cite{guidotti2018survey}. Within XAI, explainable NLP (XNLP) specifically addresses the interpretability of language-based models, focusing on features like word embeddings, attention mechanisms, and textual rationales. As Figure~\ref{fig:xnlp_pipeline} shows, a typical XNLP pipeline starts with an input text and passes it through a model. An explanation layer then makes the output easier to follow, for example by highlighting key tokens or visualizing attention weights. XNLP aims to tackle distinct linguistic challenges, such as contextual details, synonyms, or domain-specific jargon, that do not always arise in other modalities like images or tabular data~\cite{gurrapu2023rationalization, qian2021xnlp}.

Existing literature on XNLP methods, including Local Interpretable Model-Agnostic Explanations (LIME)~\cite{ribeiro_why_2016}, SHapley Additive exPlanations (SHAP)~\cite{lundberg2017unified, gramegna2021shap}, and attention-based explanations~\cite{tenney_language_2020}, has made considerable progress in unpacking model decisions. Still, much remains to be explored, especially in domain-specific contexts. In healthcare, for instance, NLP systems must provide clinically relevant insights that fit into physicians' workflows without disruption~\cite{islam2022systematic}. This necessity extends to mental health applications, where language models may assist in monitoring depressive or suicidal ideation through social media posts or electronic health records (EHRs)~\cite{zirikly2019clpsych}. In finance, explanations must address both the complexity of specialized terminology and the high stakes of compliance and fraud detection~\cite{vcernevivciene2024explainable}. Similarly, chatbots and conversational AI systems raise new challenges regarding user trust, especially if the system's responses are critical for customer support or emergency services.

%\subsection*{Motivation and Contribution}

Although general XAI frameworks~\cite{tjoa2020survey, arrieta2020explainable, puiutta2020explainable} provide broad guidelines, there is a gap in how these methods directly translate into \emph{domain-tailored} XNLP solutions. Recent surveys such as~\cite{ali2023explainable, danilevsky2020survey, islam2022systematic} argue for interpretable NLP but tend to focus on the technical aspects rather than the nuances of real-world deployment. The two standard reference points here are the post-hoc interpretability survey of~\cite{madsen2022posthoc} and the LLM explainability survey of~\cite{zhao2024explainability}. Both catalogue the available techniques thoroughly, but they organize them by mechanism rather than by the setting in which a practitioner has to decide whether to trust one. More recent reviews concentrate on LLM-centric explainability~\citep{atakishiyev2025explainability} or on multilingual XNLP~\citep{resck2025explainability}; the present survey remains the domain-organized complement to these method- and language-oriented perspectives. In response, this survey makes four main contributions.

First, we compile and examine research across seven application domains, including mental health under the medical umbrella, to show how XNLP can be adapted to different regulatory and practical requirements. As a distinctive analytical contribution, the section \emph{\nameref{subsec:cross_domain}} distills this material into a structured comparison (Table~\ref{tab:cross_domain}) of explainability requirements, methods, metrics, and challenges across the seven domains. Second, we examine the full range of evaluation approaches for XNLP from both quantitative and qualitative perspectives: we give mathematical definitions of key metrics such as fidelity and discuss human-centered measures such as user trust. We also compare the main explanation method families (Table~\ref{tab:method_comparison}) on scope, evidence of faithfulness, and computational cost. Building on that comparison, we propose a two-tier evaluation protocol that separates a shared technical core of metrics from the domain-specific validation layer through which they must be read. Third, we address open questions regarding bias, privacy, data availability, and the balance between performance and interpretability. Fourth, building on the limitations found in the existing literature, we propose research avenues such as personalized explanations, human-in-the-loop evaluation, and mechanistic interpretability for LLMs.

%\subsection*{Paper Organization}

The remainder of the paper is organized as follows. The section \emph{\nameref{sec:modeling}} reviews foundational NLP techniques and transitions to modern transformer-based methods and LLMs, highlighting their explainability mechanisms. \emph{\nameref{sec:application}} explores the application of XNLP methods in medicine (including mental health), finance, systematic reviews, CRM, chatbots and conversational AI, social and behavioral science, HR, and other emerging use cases. \emph{\nameref{sec:critical}} addresses the critical aspects of XNLP, including evaluation metrics, trade-offs, rationalization techniques, human evaluation, and data and code availability. \emph{\nameref{sec:future}} outlines promising directions for further investigation, such as personalized and mechanistic explanations. Finally, \emph{\nameref{sec:Conclusion}} summarizes the survey's key insights.

%Section 2
\section{Modeling Techniques for XNLP}
\phantomsection
\label{sec:modeling}

Figure~\ref{fig:taxonomy} summarizes the families of explanation methods reviewed in this section, organized into four broad families that recur throughout the paper: intrinsic methods, post-hoc methods, probing, and LLM-native techniques.

\begin{figure*}[ht!]
\centering
\begin{tikzpicture}[
  troot/.style={draw, thick, rectangle, align=center, font=\small\bfseries,
  fill=gray!15, inner sep=5pt},
  tfam/.style={draw, rectangle, align=center, font=\footnotesize\bfseries,
  fill=gray!5, text width=3.6cm, inner sep=4pt},
  tleaf/.style={draw, rectangle, align=left, font=\scriptsize,
  text width=3.6cm, inner sep=4pt},
  tline/.style={-stealth}]
\node[troot] (root) at (0,0) {XNLP Explanation Methods};
\node[tfam] (fa) at (-6.45,-1.4) {Intrinsic\\ (interpretable by design)};
\node[tfam] (fb) at (-2.15,-1.4) {Post-hoc};
\node[tfam] (fc) at (2.15,-1.4) {Probing};
\node[tfam] (fd) at (6.45,-1.4) {LLM-native};
\node[tleaf, below=3mm of fa] (la) {$\bullet$ Linear model coefficients (BoW, TF-IDF)\\
  $\bullet$ Attention mechanisms\\
  $\bullet$ Rationales by design};
\node[tleaf, below=3mm of fb] (lb) {$\bullet$ LIME\\
  $\bullet$ SHAP\\
  $\bullet$ Gradient-based saliency\\
  $\bullet$ Counterfactuals};
\node[tleaf, below=3mm of fc] (lc) {$\bullet$ Probing tasks for linguistic properties\\
  $\bullet$ Structural probes of representations};
\node[tleaf, below=3mm of fd] (ld) {$\bullet$ Chain-of-thought self-explanation\\
  $\bullet$ Mechanistic interpretability (sparse autoencoders)};
\draw[tline] (root) -- (fa);
\draw[tline] (root) -- (fb);
\draw[tline] (root) -- (fc);
\draw[tline] (root) -- (fd);
\draw[tline] (fa) -- (la);
\draw[tline] (fb) -- (lb);
\draw[tline] (fc) -- (lc);
\draw[tline] (fd) -- (ld);
\end{tikzpicture}
\caption{\small Taxonomy of the XNLP method families discussed in this survey: intrinsic, post-hoc, probing, and LLM-native.}
\label{fig:taxonomy}
\end{figure*}

\subsection{Explainability of Traditional NLP Models}
\label{sec:traditional_nlp_explain}

Traditional NLP models, notably Bag of Words (BoW) and its variants such as term frequency (TF) and term frequency-inverse document frequency (TF-IDF), provide foundational approaches to textual representation. A BoW model encodes a document as a set of token counts without preserving word order or contextual usage. When coupled with transparent classifiers (e.g., logistic regression), the model's coefficients help uncover each word's influence on the prediction outcome. For instance, if a logistic regression classifier assigns a large positive coefficient to the token ``excellent'' in a sentiment analysis task, it signals a strong correlation between that token and a positive sentiment~\cite{heap_word_2017, liu_representation_2020}.

TF-IDF further refines this representation by assigning greater importance to words that appear frequently in a document but are relatively rare across the entire corpus~\cite{shimomoto_text_2018, zubiaga_tf-cr_2020}. While these methods are simple and often interpretable, they struggle to encode contextual and syntactic relationships~\cite{dagan_contextual_1995, jurafsky_vector_2019, zubiaga_tf-cr_2020}. Even in TF and TF-IDF systems, the weights can mislead. When several words co-occur and jointly predict an outcome, such as disease codes in discharge letters, the model spreads the credit among them, so each receives a small coefficient. A less important word that appears on its own can end up with a larger one. Although early in conception, these classical representations paved the way for more complex models that better capture semantic relationships. The drive to incorporate contextual meaning has led researchers toward advanced embedding techniques that combine high performance with more transparent decision processes.

\subsection{Explainability of Embedding Models}
\label{sec:embedding_explain}

Embedding models revolutionized NLP by mapping words, phrases, or sentences into continuous vector spaces. In contrast to BoW-based approaches, these dense representations capture subtler semantic and syntactic details~\cite{mikolov_distributed_2013, pennington_glove_2014}. Word2Vec~\cite{mikolov_distributed_2013} and GloVe~\cite{pennington_glove_2014}, for example, produce vectors where semantically similar words (e.g., ``king'' and ``queen'') reside close together, a property earlier BoW variants could not achieve. Sentence-level embeddings (e.g., Universal Sentence Encoder~\cite{cer_universal_2018}) extend this idea by encoding entire clauses or paragraphs into fixed-dimensional vectors. Despite performance gains, these embedding models add complexity that can obscure their decision-making. Accordingly, researchers employ several explainability strategies.

The first strategy is visualization of vector spaces. Tools like TensorBoard~\cite{vogelsang2020magician} and EmbeddingVis~\cite{li2018embeddingvis} map high-dimensional embeddings into 2D or 3D layouts so that users can inspect semantic clusters and language structures. Dimensionality-reduction methods such as t-SNE~\cite{van_der_maaten_visualizing_2008} and PCA~\cite{jolliffe_principal_2016} are commonly used to reveal relationships among words or sentences~\cite{bracsoveanu2022visualizing, wang2019single}. Gradient-based saliency maps, adapted from computer vision~\cite{simonyan2013deep}, highlight the input tokens that produce the largest gradient magnitudes with respect to an embedding layer and so provide local explanations for individual predictions. Some embedding-based architectures also incorporate attention layers, which indicate how much ``focus'' the model places on particular words or sub-phrases~\cite{bahdanau2014neural, vaswani2017attention}. Unlike gradient-based methods, attention is computed during the forward pass and is available at no extra cost, though, as discussed below, attention weights should not be read directly as explanations.

Balancing interpretability and performance remains challenging as models grow increasingly complex~\cite{rudin2019we}. The notion of ``explanation'' is also context-dependent. Visualizations like PCA or t-SNE may suffice for intuitive overviews, particularly in less technical scenarios~\cite{wattenberg2016use}. However, in high-stakes domains such as medicine or finance, stakeholders often want deeper, more detailed rationales behind each prediction. Consequently, embedding-based NLP poses an ongoing tension: how to develop representations that remain accurate and powerful while offering transparent enough insights into their internal structures.

\subsection{Explainability of Transformer-Based Models}
\label{sec:transformer_explain}

Transformers have reshaped modern NLP through \emph{self-attention} mechanisms that handle long-range dependencies without recurrent networks~\cite{kalyan_ammus_2021}. A classic example is BERT~\cite{devlin2019bert}, which, by using multi-head self-attention, can be pre-trained on massive corpora and then fine-tuned for tasks like sentiment analysis, question answering, or named entity recognition~\cite{mccann2018natural}. Its successors, such as RoBERTa~\cite{liu_roberta_2019} and ALBERT~\cite{lan_albert_2020}, continue this trend, delivering improvements in various NLP benchmarks.

% \begin{figure}[t!]
%  \centering
%  \includegraphics[width=0.65\linewidth]{transformer_explanation.jpeg}
%  \caption{Illustrative depiction of transformer-based explainability. An input text is passed through multiple attention heads; attention maps can then highlight the tokens the model finds most influential when forming its final predictions. Such visualization aids in interpreting how different parts of the input interact~\cite{vig2019multiscale}.}
%  \label{fig:transformer_expl}
% \end{figure}

\begin{figure}[t!]
  \centering
  \begin{subfigure}[b]{\linewidth}
  \includegraphics[width=\linewidth]{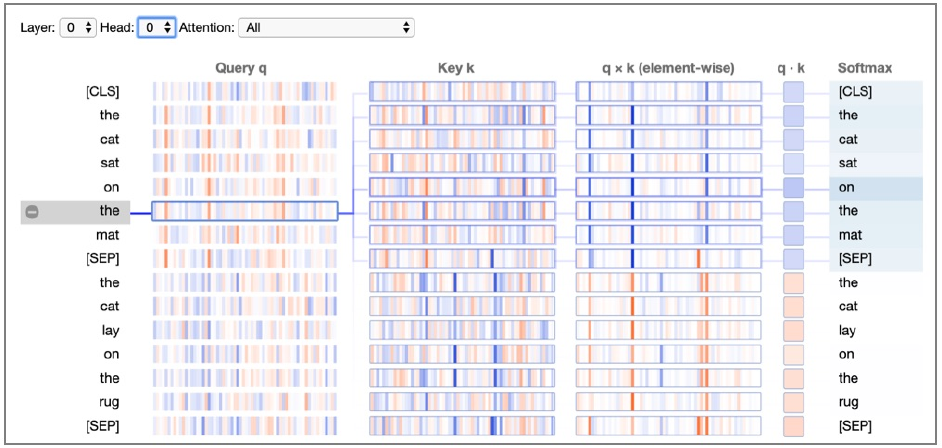}
  \caption{\small BERT (bidirectional encoder)}
  \label{fig:neuron_bert}
  \end{subfigure}
  \par\medskip
  \begin{subfigure}[b]{\linewidth}
  \includegraphics[width=\linewidth]{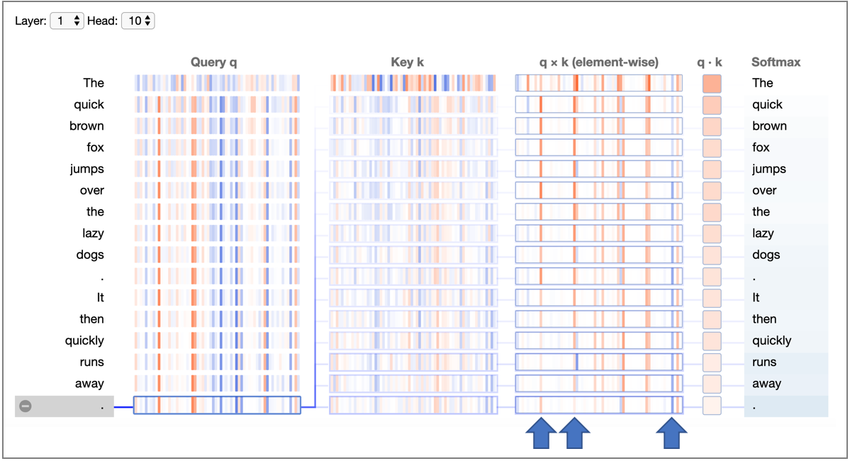}
  \caption{\small GPT-2 (autoregressive decoder), layer 1, head 10}
  \label{fig:neuron_gpt2}
  \end{subfigure}
  \caption{Neuron view of BERTViz~\citep{vig2019multiscale} for single attention heads. In BERT (a), a token can attend to context on both sides; in GPT-2 (b), causal masking limits attention to preceding tokens.}
  \label{fig:transformer_expl}
\end{figure}

Although Transformer architectures achieve strong results on NLP tasks, they present unique interpretability challenges. A widely adopted strategy is attention-weight visualization, as illustrated in Figure~\ref{fig:transformer_expl} for both an encoder-based model (BERT) and a decoder-based model (GPT-2). Tools like BERTViz enable multiscale inspection of attention patterns across layers and heads, helping researchers and practitioners understand which input tokens the model attends to when generating its predictions~\cite{vig2019multiscale}. Comparing the two panels also makes an architectural difference directly visible: the bidirectional encoder can distribute attention across the full sentence, while the autoregressive decoder attends only to earlier positions. This form of visualization helps in interpreting token interactions and can reveal model reasoning for specific linguistic phenomena. Several other method families help unpack transformer-based decisions.

Probing tasks are simplified linguistic evaluations that reveal which grammatical or semantic properties the model retains~\cite{hewitt2019designing}. Feature visualization depicts learned features or activation patterns, showing which elements particular words or phrases activate~\cite{olah2017feature}. Attribution methods such as Integrated Gradients~\cite{sundararajan2017axiomatic} assign token-level importance scores by accumulating gradients over input perturbations. Model simplification and distillation train smaller, more interpretable student networks that replicate a large model's outputs, bridging the gap between high performance and clarity~\cite{hinton2015distilling}.

Model-agnostic attribution methods are also frequently integrated with transformers. Both LIME~\cite{ribeiro_why_2016} and SHAP~\cite{lundberg2017unified} attribute a single prediction to individual input features; a global picture emerges only when such local attributions are aggregated over a corpus~\cite{ribeiro2016model}. Perturbation-based techniques~\cite{liang_deep_2018} systematically alter input text to identify critical words or phrases, while contextual decomposition~\cite{murdoch_beyond_2018} breaks down output scores into contributions from individual tokens or token interactions. Despite the proliferation of these tools, no single best practice exists for all contexts. A further caveat is that attention-weight visualization can mislead: in sentiment analysis, weights could highlight neutral tokens (e.g., ``the''), downplaying salient words like ``hate''~\cite{jain_attention_2019, wiegreffe_attention_2019}. As a result, any claim of interpretability demands careful validation against the model's actual decision-making mechanisms and the end-user's needs. 

Transformer models like BERT~\cite{devlin2019bert} and GPT-x~\cite{radford2019language} have led to major advances in NLP, but their large size and complex inner workings make it hard to understand how they work. The section \emph{\nameref{sec:critical}} discusses many of these explainability challenges in practice. We also expect that future XNLP methods will include better built-in ways to make their decisions easier to understand.

\subsection{LLMs and Mechanistic Interpretability}
\label{sec:llm_mechanistic}

Building on transformer foundations, recent LLMs like Claude~3, GPT-4o, and Gemini have driven major advances in interpretability research, often called \emph{mechanistic} or \emph{feature-level} interpretability. Unlike post-hoc methods that explain model outputs after the fact, mechanistic interpretability seeks to reverse-engineer the internal computations of neural networks, identifying the specific circuits and features responsible for particular behaviors~\citep{templeton2024scaling}.

\subsubsection{Sparse Autoencoders: Progress, Limitations, and Alternatives}

A central challenge in understanding LLMs is \emph{polysemanticity}: individual neurons often encode multiple unrelated concepts, making interpretation difficult~\citep{cunningham2024sparse}. Sparse autoencoders (SAEs) address this by learning to decompose neural activations into sparse, interpretable features. The key insight is that while individual neurons may be polysemantic, the underlying representations can be disentangled into monosemantic directions in activation space.

Anthropic's landmark work on Claude~3 Sonnet~\citep{templeton2024scaling} demonstrated that SAEs can extract millions of interpretable features from LLM activations. Their analysis surfaced features spanning both concrete entities, such as the Golden Gate Bridge, and abstract notions such as sycophancy and deception, and showed that the same features generalize across languages and to images despite text-only training. Importantly, these features are not merely correlational; intervening on specific features (e.g., amplifying the ``Golden Gate Bridge'' feature) systematically altered model outputs in predictable ways.

Recent architectural improvements have enhanced SAE effectiveness. BatchTopK SAEs~\citep{bussmann2024batchtopk} improve training efficiency by applying top-k sparsity across batches rather than individual samples, achieving better reconstruction fidelity with fewer active features. Similarly, \emph{transcoders}~\citep{dunefsky2024transcoders} extend the idea to whole MLP sublayers, approximating a dense layer with a wider sparsely activating one and making circuit analysis through those sublayers tractable. Building on transcoders, \emph{attribution graphs} enable circuit tracing, reconstructing the intermediate computational steps a model performs on individual prompts~\citep{lindsey2025biology}.

Despite promising results, mechanistic interpretability faces serious challenges. Recent critiques~\citep{chanin2024saeconceptdetection} have shown that SAEs may fail to reliably detect concepts they purportedly encode, raising questions about whether extracted features genuinely represent causal mechanisms or merely correlate with interpretable concepts. Head-to-head evaluations further show that steering model behavior through SAE features underperforms simpler prompting and finetuning baselines~\citep{wu2025axbench}. The scalability of these methods to frontier models with hundreds of billions of parameters remains uncertain, and the computational cost of training SAEs on large models is substantial.

Parallel techniques complement SAE-based approaches. B-cosification~\citep{arya2024b} retrofits transformers with constrained linear layers so that each weight contributes a signed, additive explanation of the output, advocating for \emph{explainability-by-design}. Explanation-distillation methods~\citep{fang2025knowledge} train smaller student models to replicate both task outputs \emph{and} rationales of LLM teachers, enabling compact models faithful to original reasoning.

\subsubsection{The Chain-of-Thought Faithfulness Problem}
\label{subsec:cot_faithfulness}

A critical question for LLM explainability is whether Chain-of-Thought (CoT) reasoning~\citep{wei2022chain} provides faithful explanations of model behavior. CoT prompting elicits step-by-step reasoning from LLMs, seemingly showing their reasoning process. However, emerging evidence suggests that CoT outputs may not accurately reflect the model's actual computational process, a phenomenon termed the \emph{faithfulness problem}~\citep{barez2025cot}. Turpin et al.~\cite{turpin2023language} demonstrated through intervention studies that LLMs frequently produce plausible-sounding reasoning that does not correspond to their true decision-making mechanisms. They introduced biasing features into prompts and found that models often gave correct answers while providing reasoning chains that ignored the biasing information entirely, an ``illusion of transparency.'' More concerningly, Lanham et al.~\cite{lanham2023measuring} found that larger models exhibited \emph{less} faithful reasoning than smaller ones, suggesting that increased capability may worsen rather than improve the faithfulness problem. Their study also operationalizes faithfulness through a battery of quantitative probes (early answering, adding mistakes, truncating the chain, and paraphrasing it) that test whether the final answer actually depends on the stated reasoning~\citep{lanham2023measuring}. Recent evidence extends these concerns to dedicated reasoning models, which verbalize hints they demonstrably rely on in under 20\% of probed cases~\citep{chen2025reasoning}, and shows that unfaithful reasoning, including implicit post-hoc rationalization, arises even on unperturbed prompts~\citep{arcuschin2026chain}.

These findings have important implications for XNLP. If CoT explanations do not faithfully represent model reasoning, they cannot serve as reliable explanations for high-stakes decisions. Several remedies have been proposed. Lyu et al.~\cite{lyu2023faithful} separate reasoning from computation: the LLM first translates the problem into a formal representation, and a deterministic solver then produces the final answer, so the stated reasoning chain is faithful to the computation by construction. Intervention-based validation instead perturbs CoT steps systematically and measures the effect on outputs, distinguishing faithful from unfaithful reasoning~\citep{turpin2023language}. A third direction grounds CoT mechanistically: tracing which internal features are active while a model reasons may provide stronger faithfulness guarantees than inspecting the verbalized chain alone~\citep{dunefsky2024transcoders, lindsey2025biology}.

The CoT faithfulness problem points to a broader tension in XNLP: the difference between \emph{plausible} explanations that satisfy users and \emph{faithful} explanations that accurately represent model behavior. As LLMs increasingly generate their own explanations, validating faithfulness becomes the central problem for trustworthy AI deployment.

\subsubsection{Toward Explainability-by-Design}

This body of research points to a shift toward deeper transparency in LLMs. Instead of relying solely on post-hoc methods like attention or gradient-based saliency, researchers increasingly advocate embedding interpretability directly into model architectures and training, a case argued forcefully for high-stakes decisions by~\cite{rudin2019stop}. Recent work on weight-sparse transformers exemplifies this direction, training models whose internal circuits are interpretable by construction~\citep{gao2025weightsparse}. Readers seeking a consolidated entry point to this literature can turn to the primer on transformer internals by~\cite{ferrando2024primer} and the review of mechanistic interpretability for AI safety by~\cite{bereska2024mechanistic}. As LLMs grow in complexity and capability, mechanistic interpretability and faithful reasoning validation become central to keeping these systems accountable, trustworthy, and aligned with human values~\citep{keles2024transformer, sun2025stable}.

%Secton 3
\section{Applications and Domains of XNLP}
\phantomsection
\label{sec:application}

AI-based applications have been extensively adopted across multiple facets of modern life, including social media, medicine, commerce, customer service, and finance. Common tasks like machine translation, text summarization, and sentiment analysis aim to automate routine processes and improve user experiences~\cite{vaswani2017attention, zhao_survey_2023}. In these generic contexts, performance often takes precedence over the transparency of how a model arrives at its conclusions. However, in more sensitive fields such as medicine, or in high-stakes areas like finance, achieving explainability is critical. For instance, clinicians analyzing EHRs must trust not only the accuracy but also the reasoning behind a model's predictions, given that such insights directly influence patient treatment plans. A prediction on its own answers only \emph{what} the model concluded; XNLP additionally seeks to answer \emph{why} a particular output was produced, \emph{which} parts of the input drove it, and \emph{when} the system should not be relied upon. These rationales must be clear and actionable so that healthcare providers, financial analysts, and other key stakeholders can have confidence in the system.

Beyond interpretability, fairness and equity rank highly in domains where decisions can drastically impact individuals or communities~\cite{barocas2016big}. In medicine, for example, demonstrating how a patient is flagged for a specific intervention is essential to ensuring unbiased healthcare. Similarly, in CRM, AI-driven personalization should not discriminate based on sensitive attributes. As shown by Brandl et al.~\cite{brandl-etal-2024-interplay}, balancing fairness and explainability can promote greater trust in AI systems. This aligns with the principle that if logical and scientifically grounded arguments reinforce current knowledge, users are more inclined to trust an AI's conclusions~\cite{islam2021local}.

XNLP also benefits human-AI collaboration. When an AI exceeds human capability in certain tasks, such as strategic game play, e.g., AlphaGo~\cite{silver2016mastering}, transparent explanations allow individuals to glean novel strategies or insights. Conversely, if an AI achieves performance comparable to a human expert, interpretability reinforces end-user trust in the system~\cite{glikson2020human}. In finance, for example, a model predicting firm valuations must provide transparent justifications so that shareholders can audit its outputs. Absent transparency, the model could be manipulated to favor particular interests. Similarly, chatbots in customer service must justify their responses, not merely produce them, if they are to earn user trust and satisfaction. 

Table~\ref{tab:applications} offers an overview of some of the key application domains where XNLP has had substantial practical impact. These domains range from medicine and finance to systematic reviews, CRM, chatbots, and social and behavioral science, each with distinct subcategories and challenges. HR is also included as a growing domain where XNLP can augment processes like recruitment and performance evaluation. The section \emph{\nameref{subsec:cross_domain}} and Table~\ref{tab:cross_domain} synthesize how these requirements diverge across domains.

\begin{table*}[ht!]
\centering
\scriptsize
\renewcommand{\arraystretch}{1}
\caption{Overview of XNLP Applications, Subcategories, and Case Studies}
\label{tab:applications}
\begin{tabular}{p{3.5cm} p{5.5cm} p{3.2cm}}
\toprule
\textbf{XNLP Applications} & \textbf{Subcategories} & \textbf{Studies} \\
\midrule

Medicine & EHRs 
&~\cite{choi_doctor_2016, alabi2023machine} \\
& Medical Documents Analysis 
&~\cite{li_neural_2022, moradi_deep_2022, kang_eliie_2017, weng_medical_2017} \\
\addlinespace

Finance & Risk Assessment
&~\cite{fritz-morgenthal_financial_2022} \\
& Fraud Detection
&~\cite{psychoula_explainable_2021, fritz-morgenthal_financial_2022, cirqueira_towards_2021} \\
& Firm Valuation 
&~\cite{rizinski2024sentiment, bagga2023towards} \\
\addlinespace

Systematic Reviews & Review Automation
&~\cite{ferdinands2023active} \\
& Bias Assessment
&~\cite{marshall_robotreviewer_2016} \\
\addlinespace

CRM & Sentiment Analysis 
&~\cite{capuano2021sentiment, bacco_explainable_2021} \\
& Customer Support Automation
&~\cite{jenneboer2022impact, guerreiro2024systematic} \\
\addlinespace

Chatbots and Conversational AI & Conversational Agents
&~\cite{ghazvininejad2018knowledge, kovari2025explainable} \\
& Context-Aware Recommendations 
&~\cite{zhang_deep_2019} \\
\addlinespace

Social and Behavioral Science & Sexism and Hate Speech Detection 
&~\cite{mozafari2020hate, saleh2023detection, plaza2023overview, kirk2023semeval, mohammadi2023towards, mohammadi2024transparent, anjum2023hate} \\
& Fake News and AI Generative Detection 
&~\cite{capuano2023content, mohammadi_2023_10079010} \\
\addlinespace

Human Resources & Talent Acquisition and Recruitment 
&~\cite{patwardhan2023transformers, gurrapu2023rationalization} \\
& Employee Sentiment Analysis 
&~\cite{jim2024recent, danilevsky2020survey} \\
& Performance Evaluation 
&~\cite{chang2023natural, herrewijnen2023human} \\
& Bias Auditing in Recruitment
&~\cite{kaushal2024resume, webster2025fairness} \\

\bottomrule
\end{tabular}
\end{table*}

\subsection{Medicine}
\label{sec:medicine}

XNLP has become particularly valuable in the medical domain, where decisions can be life-altering. From analyzing patient histories in EHRs to processing physician notes for disease risk assessments, explainability is key because it clarifies the logic behind a model's prediction. For example, Choi et al.~\cite{choi_retain_2016} devised an interpretable recurrent neural network (RNN) model for heart failure prediction, explicitly highlighting which medical codes contributed most to the risk factors. Similar rule-based approaches have been used to extract clinical evidence from randomized controlled trials~\cite{kang_eliie_2017}, ensuring transparency in processes that directly influence patient care.

Recent initiatives also extend to mental health, a critical area of medicine that relies on sensitive textual data, often collected from social media platforms or patient narratives. For instance, LLM-based analyses have been proposed to detect early signs of depression or suicidal ideation, using model explanations to reassure clinicians and researchers about the validity of identified risk factors~\cite{yang-etal-2023-towards}.

These complementary explanation methods can be illustrated through heart failure prediction. The RETAIN model~\citep{choi_retain_2016} was trained on 14 million visits from 263,000 patients and reaches an AUC of 0.8705, on par with an unconstrained RNN baseline at 0.8706. It stays interpretable through a two-level attention mechanism. Visit-level weights ($\alpha$) show which clinical encounters mattered most to the prediction, and code-level weights ($\beta$) pick out the diagnoses and medications within each visit. For a patient flagged as high-risk, RETAIN might reveal that the model attended primarily to a recent hospitalization (e.g., visit attention $\alpha = 0.31$), with diagnosis codes for cardiac dysrhythmia and heart failure receiving the highest code-level contributions ($\beta > 0.15$). Applying SHAP~\citep{lundberg2017unified} to the same prediction produces additive feature attributions, where studies show that serum creatinine levels typically emerge as the strongest predictor, with elevated values contributing positively to mortality risk while normal values reduce predicted risk~\citep{miao2023interpretable}. Meanwhile, LIME~\citep{ribeiro_why_2016} fits a local linear approximation, enabling counterfactual reasoning such as estimating how risk would change if a specific diagnosis were absent. Each method illuminates different aspects: RETAIN reveals temporal patterns through reverse-time attention, SHAP provides globally consistent feature attributions grounded in game theory, and LIME offers locally faithful approximations for individual predictions. Recent meta-analyses of XAI in clinical decision support systems~\citep{nazir2025xai, noor2025unveiling} emphasize that combining multiple explanation methods provides clinicians with richer decision support than any single method alone.

Table~\ref{tab:summary1} provides a concise overview of major XNLP applications in medicine, showing the interplay among model architectures, explainability techniques, and evaluation metrics.

% \begin{table*}[ht!]
% \centering
% \scriptsize
% \renewcommand{\arraystretch}{0.9}
% \caption{Summary of XNLP Applications in Medicine.}
% \label{tab:summary1}
% \begin{tabular}{p{1.5cm} p{2.2cm} p{2.2cm} p{3cm} p{2cm} p{1.8cm}}
% \toprule
% \textbf{Paper} & \textbf{Application} & \textbf{Model} & \textbf{Explainability Method} & \textbf{Dataset} & \textbf{Metrics} \\ 
% \midrule

%~\cite{choi_doctor_2016}
% & Heart Failure Prediction 
% & RNN 
% & Feature Importance 
% & EHRs 
% & Accuracy, AUC-ROC \\
% \addlinespace

%~\cite{mullenbach2018explainable}
% & EHRs Classification 
% & CNN 
% & Rationale-based explanations 
% & MIMIC-III 
% & Precision, Recall, F1-score \\
% \addlinespace

%~\cite{kang_eliie_2017} 
% & RCT Analysis \& Extraction 
% & Rule-based NLP 
% & Transparent Rule Extraction 
% & Clinical Trials 
% & Extraction Accuracy \\
% \addlinespace

%~\cite{zhang2019biowordvec}
% & Biomedical Word Embeddings 
% & fastText + MeSH 
% & MeSH-Based Interpretations 
% & UMNSRS-Sim, UMNSRS-Rel 
% & Embedding Quality \\
% \addlinespace

%~\cite{yang2021deep}
% & Disease Progression Prediction 
% & Transformer 
% & Attention Mechanism 
% & Public EHRs 
% & RMSE, MAE \\
% \addlinespace

%~\cite{alabi2023machine}
% & Cancer Diagnosis 
% & BERT 
% & LIME and SHAP 
% & Cancer Registry 
% & F1-score, Precision, Recall \\

% \bottomrule
% \end{tabular}
% \end{table*}

%%%%%%%%%%%%%%%%%%%%%%%%%%%%%%%%%%%%%%%%%%%%%%%%%%%%%%%%
% 1) XNLP Applications in Medicine
%%%%%%%%%%%%%%%%%%%%%%%%%%%%%%%%%%%%%%%%%%%%%%%%%%%%%%%%
\begin{table}[ht!]
\centering
\scriptsize
\renewcommand{\arraystretch}{0.9}
\caption{Summary of XNLP Applications in Medicine}
\label{tab:summary1}
\setlength{\tabcolsep}{4pt}%
\begin{tabular}{P{0.55cm} P{1.3cm} P{1.1cm} P{1.3cm} P{1.0cm} P{1.1cm}}
\toprule
\textbf{Ref.} & \textbf{Application} & \textbf{Model} 
& \textbf{Explainability Method} & \textbf{Dataset} & \textbf{Metrics} \\ 
\midrule

\cite{choi_retain_2016}
& Heart Failure Prediction
& RNN
& Attention Mechanism
& EHRs
& Accuracy, AUC-ROC \\

\addlinespace
\cite{mullenbach2018explainable}
& EHRs Classification 
& CNN 
& Rationale-based explanations 
& MIMIC-III 
& Precision, Recall, F1-score \\

\addlinespace
\cite{kang_eliie_2017}
& RCT Analysis \& Extraction 
& Rule-based NLP 
& Transparent Rule Extraction 
& Clinical Trials 
& Extraction Accuracy \\

\addlinespace
\cite{zhang2019biowordvec}
& Biomedical Word Embeddings 
& fastText + MeSH 
& MeSH-Based Interpretations 
& UMNSRS-Sim, UMNSRS-Rel 
& Embedding Quality \\

\addlinespace
\cite{yang2021deep}
& Disease Progression Prediction
& Deep multimodal model
& Attention Mechanism
& Multimodal clinical data
& RMSE, MAE \\

\addlinespace
\cite{alabi2023machine}
& Cancer Diagnosis 
& BERT 
& LIME and SHAP 
& Cancer Registry 
& F1-score, Precision, Recall \\

\bottomrule
\end{tabular}
\end{table}

Alongside EHR classification and rule-based extraction, contemporary work uses deep learning methods, e.g., word embeddings, transformers, to produce stronger predictive power in domains like cancer diagnostics~\cite{alabi2023machine}, disease progression modeling~\cite{yang2021deep}, and automated ICD coding~\cite{mullenbach2018explainable}. Nonetheless, the need for interpretability remains pressing, as stakeholders require transparent models to validate medical recommendations and address concerns about potential biases. XAI enhances trust among healthcare professionals by explaining AI-driven decisions, thereby meeting regulatory transparency requirements and promoting fairness and safety in clinical settings~\cite{abgrall2024should}.

Research on mental health analysis via XNLP further illustrates the domain's complexity and sensitivity. For instance, LLM-based strategies to detect distress patterns depend on clear, faithful explanations that can be passed to mental health practitioners~\cite{yang-etal-2023-towards}. Ensuring patient privacy, dealing with text de-identification~\cite{Bagheri2023}, and mitigating data biases remain the main obstacles in this space. Overall, integrating XNLP in the medical domain holds the promise of safer clinical decision support, faster literature reviews, and more equitable patient care by revealing how and why a system recommends certain treatments or diagnoses.

\noindent
\textbf{Domain-specific challenges.} Deploying XNLP in medicine faces obstacles beyond model design: HIPAA/GDPR privacy constraints restrict data sharing, concept drift in longitudinal EHR data erodes both accuracy and explanation validity, explanations must fit into time-pressured clinical workflows, and unresolved liability questions surround AI-informed decisions~\citep{nazir2025xai, noor2025unveiling}. Recent work begins to address the evaluation side of these challenges: benchmarks pairing challenging medical questions with expert-written explanations enable quantitative assessment of LLM-generated medical rationales~\citep{chen2025benchmarking}, and clinician studies show that the format of an explanation, whether SHAP visualizations or clinician-friendly narratives, measurably changes clinical decision behavior~\citep{hur2025comparison}.

\subsection{Finance}
\label{sec:finance}

Financial decision-making involves complex processes such as risk assessment, fraud detection, and firm valuation, all of which require both accurate and transparent AI-driven insights. Recent studies show that incorporating NLP techniques to analyze financial reports, news articles, or transaction logs can effectively flag risk factors and provide timely alerts~\cite{wang2024application}. However, traditional NLP-based models often lack explainability, which is critical when end users, be they financial experts or customers, need to understand why a model has flagged a transaction as fraudulent or assigned a particular credit score. XAI addresses this gap by providing interpretable insights that can build trust in the system's outputs. For instance, integrating XAI techniques into credit scoring models enhances transparency and compliance with regulations like the General Data Protection Regulation (GDPR) and the Equal Credit Opportunity Act (ECOA), ensuring that algorithmic decisions are understandable and coherent~\cite{demajo2020explainable}. Employing model-agnostic explanation methods such as SHAP and LIME in credit risk management also helps stakeholders understand the reasoning behind model predictions, which supports trust and better-informed decision-making~\cite{misheva2021explainable}.

%\subsubsection*{Risk Assessment}
Risk assessment underpins core tasks in the financial sector, such as loan approvals, insurance rate settings, and investment decisions. Inaccuracies or lack of transparency in this process can have serious consequences, from unfair interest rates to systemic risks. XNLP can help by showing which textual factors, like specific keywords in a credit application or trends in financial reports, contribute to elevated risk scores~\cite{fritz-morgenthal_financial_2022}. For example, Demajo et al.~\cite{demajo2020explainable} combined credit scoring models with global and local explanation techniques, highlighting how individual features shape the final score. Such clarity serves two audiences: it justifies decisions to regulators and auditors, and it shows clients what they would need to change to lower their risk~\cite{rudin2022interpretable}. A recent PRISMA-based review of LLM-based credit scoring taxonomizes the explainability mechanisms employed across the domain~\citep{golec2025interpretable}. Caution is warranted, however: for loan-default prediction, LLM self-explanations have been shown to diverge from empirical SHAP attributions of the same predictions~\citep{almarri2025interpreting}.

%\subsubsection*{Fraud Detection}
Fraud detection systems have traditionally operated as opaque ``black boxes,'' leaving end users uncertain about what triggers a fraud alert. XNLP tools shine a light on relevant textual features, possibly certain transaction notes or unusual patterns in communications, to clarify why a specific transaction has been flagged~\cite{psychoula_explainable_2021, fritz-morgenthal_financial_2022, cirqueira_towards_2021}. Interpretable machine learning frameworks, such as decision trees, can provide feature-importance scores, while neural architectures can integrate attention layers that visually highlight suspicious keywords. For instance, Cirqueira et al.~\cite{cirqueira_towards_2021} introduced an XAI approach for fraud detection, aligning fraud experts' investigative tasks with model-generated explanations. Better comprehension of these flags reduces false positives and strengthens collaboration between human fraud analysts and AI systems.

The practical value of combining ensemble methods with explainability is well demonstrated in fraud detection research~\citep{hassan2025fraud, hassan2025model}. A stacking ensemble of XGBoost, LightGBM, and CatBoost evaluated on the IEEE-CIS Fraud Detection dataset, which contains more than 590,000 real transaction records, reaches 99\% accuracy and an AUC-ROC of 0.99~\citep{hassan2025fraud}. SHAP is used there for feature selection, identifying the most influential predictors, with LIME, partial dependence plots, and permutation feature importance layered on top to explain individual decisions. When such a model flags a transaction, SHAP waterfall plots break the prediction into individual feature contributions. An unusually high amount for that cardholder pushes the fraud score up, while a long account history and a consistent merchant category pull it down. Exploratory analyses further reveal that most fraudulent transactions involve amounts below \$1,000, exhibiting more dispersed value distributions compared to legitimate transactions~\citep{hassan2025fraud}. This detailed breakdown lets fraud analysts verify whether flagged features align with known fraud patterns, improving investigative efficiency. However, Khan et al.~\cite{hassan2025model} note that model-agnostic methods like SHAP face scalability challenges in real-time systems processing thousands of transactions per second, motivating research into approximate and cached explanation methods that trade some precision for computational efficiency~\citep{vcernevivciene2024explainable}.

%\subsubsection*{Firm Valuation}
XNLP methods are also gaining traction in firm valuation, a high-stakes arena of finance where annual reports, market data, and corporate disclosures must be parsed to determine a company's worth. A key challenge lies in distinguishing relevant signals from strategic or even misleading language. More advanced NLP models, including transformer-based architectures, attempt to capture long-range dependencies and contextual nuances~\cite{bagga2023towards}. Yet context-dependent meanings, sarcasm, and subtle cues remain challenging to detect, raising concerns that companies might ``game the system'' by inserting specific words likely to inflate valuations~\cite{abro2023natural}. To address this, XNLP solutions like the one proposed by Bagga and Stathis~\cite{bagga2023towards} incorporate interpretability mechanisms (e.g., attention-based explanations) to show precisely which textual segments influenced a model's valuation output. Similarly, Rizinski et al.~\cite{rizinski2024sentiment} found that combining explainable lexicon models with sentiment analysis in financial texts improves both accuracy and interpretability, allowing investors and auditors to discern exactly how sentiment-laden phrases affect overall firm valuation.

\begin{table}[ht!]
\centering
\scriptsize
\renewcommand{\arraystretch}{0.9}
\caption{Summary of XNLP Applications in Finance}
\label{tab:summary2}
\setlength{\tabcolsep}{4pt}%
\begin{tabular}{P{0.55cm} P{1.3cm} P{1.1cm} P{1.3cm} P{1.0cm} P{1.1cm}}
\toprule
\textbf{Ref.} & \textbf{Application} & \textbf{Model} 
& \textbf{Explainability Method} & \textbf{Dataset} & \textbf{Metrics} \\ 
\midrule

\cite{cirqueira_towards_2021}
& Fraudulent Transactions Justification 
& XAI Methods 
& Explanation Generation 
& Financial Transactions 
& Accuracy, F1-score, Precision, Recall \\

\addlinespace
\cite{bagga2023towards}
& Firm Valuation 
& Transformer-based Model 
& Explanation Generation 
& Financial Documents 
& ROUGE, BLEU \\

\addlinespace
\cite{rizinski2024sentiment}
& Sentiment Analysis for Valuation 
& Explainable Lexicon Model 
& SHAP Explainability 
& Financial Texts 
& Accuracy, F1-score, Precision, Recall \\

\bottomrule
\end{tabular}
\end{table}

\noindent
In summary, XNLP is reshaping key finance processes by injecting interpretability into risk analyses, fraud alerts, and valuation metrics. Explainable risk assessments help users manage their creditworthiness more effectively, while transparent fraud detection can strengthen the reliability and acceptance of AI-driven alerts. Meanwhile, interpretable firm valuation models demystify how textual content in financial disclosures influences market perceptions. As Table~\ref{tab:summary2} illustrates, the state of the art spans various architectures (from decision trees to transformers) and explanation frameworks (from feature-importance scores to saliency maps), reflecting a rapidly evolving field. Going forward, deeper integration of XNLP principles, coupled with sound evaluation metrics, will be central to boosting stakeholder confidence and ensuring more equitable, auditable financial ecosystems.

\subsection{Systematic Reviews}
\label{sec:systematic_reviews}

Systematic reviews are a cornerstone of evidence-based decision-making in various fields. The process commonly involves screening large volumes of research literature, extracting relevant data, and synthesizing key findings. Although AI-based automation can accelerate tasks such as study identification or data extraction, explainability determines whether researchers can see why specific studies were included or excluded. Techniques derived from XNLP help demystify this process, offering transparency and trust in the automated workflow.

For instance, Rosemblat et al.~\cite{rosemblat_towards_2019} use semantic predications from SemMedDB to detect and characterize apparent contradictions in the biomedical literature, making conflicting evidence more transparent to reviewers. Such methods, collectively, allow reviewers to focus on interpreting potentially relevant papers rather than sifting through thousands of irrelevant ones.

%~\cite{van_de_schoot_open_2021} introduced ASReview\footnote{\scriptsize\url{https://asreview.nl/}}, an open-source active learning platform designed to streamline the study selection phase of systematic reviews. By integrating explanation features, ASReview provides users with insight into why certain articles are flagged as relevant, thus enabling them to audit and refine the model's decisions. Meanwhile

Marshall et al.~\cite{marshall_robotreviewer_2016} developed RobotReviewer, which addresses bias assessment within systematic reviews. By using text analysis for bias detection in clinical trials, RobotReviewer can automatically surface key phrases or patterns indicative of methodological flaws. When combined with an explainable component, this process speeds up bias assessment and clarifies the reasons behind each flagged instance. Table~\ref{tab:summary3} highlights notable XNLP applications in systematic reviews, detailing the associated models, explainability methods, data sources, and relevant metrics.

% \begin{scriptsize}
% \centering
% \renewcommand{\arraystretch}{0.9}
% \begin{longtable}{p{0.2cm} p{0.1cm} p{1.7cm} p{2.5cm} p{2.2cm} p{1.9cm} p{2.2cm}}
% \caption{Summary of XNLP Applications in Systematic Reviews} \label{tab:summary3} \\
% \toprule
% \textbf{Paper} & \textbf{} & \textbf{Application} & \textbf{Model} & \textbf{Explainability Method} & \textbf{Dataset} & \textbf{Metrics} \\ 
% \midrule
% \endfirsthead
% \toprule
% \textbf{Paper} & \textbf{} & \textbf{Application} & \textbf{Model} & \textbf{Explainability Method} & \textbf{Dataset} & \textbf{Metrics} \\ 
% \midrule
% \endhead

%~\cite{rosemblat_towards_2019} & & Review Automation 
% & Support Vector Machine (SVM) 
% & Explanation Framework 
% & PubMed 
% & WSS\footnote{\scriptsize Work Saved over Sampling}, RRF\footnote{\scriptsize Relevant References Found} \\ 
% \addlinespace

%~\cite{marshall_robotreviewer_2016} & & Bias Assessment 
% & RobotReviewer 
% & Text Analysis for Bias Detection 
% & Clinical Trials 
% & Bias Assessment Metrics, Accuracy \\ 
% \addlinespace

%~\cite{chen2018learning} & & Model Interpretation 
% & LSTM 
% & Information Bottleneck 
% & SST-2, IMDb 
% & Accuracy, Mutual Information \\ 
% \bottomrule
% \end{longtable}
% \end{scriptsize}

%%%%%%%%%%%%%%%%%%%%%%%%%%%%%%%%%%%%%%%%%%%%%%%%%%%%%%%%
% 3) XNLP Applications in Systematic Reviews
%%%%%%%%%%%%%%%%%%%%%%%%%%%%%%%%%%%%%%%%%%%%%%%%%%%%%%%%
\begin{table}[ht!]
\centering
\scriptsize
\renewcommand{\arraystretch}{0.9}
\caption{Summary of XNLP Applications in Systematic Reviews}
\label{tab:summary3}
\setlength{\tabcolsep}{4pt}%
\begin{tabular}{P{0.55cm} P{1.3cm} P{1.1cm} P{1.3cm} P{1.0cm} P{1.1cm}}
\toprule
\textbf{Ref.} & \textbf{Application} & \textbf{Model} 
& \textbf{Explainability Method} & \textbf{Dataset} & \textbf{Metrics} \\ 
\midrule

\cite{rosemblat_towards_2019}
& Contradiction Detection in Literature
& Rule-based NLP (Semantic Predications)
& SemMedDB Predication Analysis
& PubMed
& Manual Review \\

\addlinespace
\cite{marshall_robotreviewer_2016}
& Bias Assessment 
& RobotReviewer 
& Text Analysis for Bias Detection 
& Clinical Trials 
& Bias Assessment Metrics, Accuracy \\

\bottomrule
\end{tabular}
\end{table}

Recent large-scale simulation studies have shown that active learning models enhance the efficiency of systematic review screening processes. By prioritizing the most relevant records, these models reduce the manual workload required for reviewing literature. For instance, Teijema et al.~\cite{teijema2025large} conducted over 29,000 simulations across various model configurations and datasets, consistently finding that active learning outperformed random screening strategies in identifying relevant studies. A scoping review of 48 such simulation studies, spanning 208 datasets and 15 model configurations, consolidates this quantitative evidence on active-learning screening efficiency~\citep{teijema2025simulation}.

Quantitative evaluations demonstrate the substantial efficiency gains achievable through these approaches. Simulation studies comparing classifiers such as SVMs with TF-IDF feature extraction show that active learning substantially shortens the Average Time to Discovery (ATD) of relevant records relative to random screening~\citep{ferdinands2023active}. In food safety literature screening, active learning models achieved 99.2\% recall while reviewing only 62.6\% of total records~\citep{burgard2025active}. For bias assessment in clinical trials, RobotReviewer demonstrates inter-rater reliability (Cohen's $\kappa$) of 0.42 to 0.48 across risk-of-bias domains, comparable to agreement levels between human reviewers~\citep{marshall_robotreviewer_2016}. These metrics show that XNLP tools can match human-level performance while dramatically reducing screening burden, though the trade-off between recall and efficiency requires careful calibration based on review objectives. Moving beyond screening alone, an end-to-end LLM workflow benchmarked on 32,357 citations reported higher screening sensitivity than human reviewers (96.7\% versus 81.7\%), extending automation to data extraction and bias assessment~\citep{cao2025automation}. As the volume of global research continues to grow, incorporating XNLP into automated reviewing workflows stands as a promising strategy to enhance speed, clarity, and consistency in evidence synthesis.

\subsection{Customer Relationship Management (CRM)}
\label{sec:crm}

CRM involves diverse functions such as tracking user sentiment, generating automated summaries of user feedback, and providing responsive customer support. Recent advances in XNLP have shown promise in improving the transparency and effectiveness of these processes. By revealing the underlying factors that drive model outputs, organizations can better trust, refine, and act upon AI-generated insights.

A key role for XNLP in CRM emerges in text summarization, where textual data from various sources, such as product reviews or user comments, must be condensed into concise, informative summaries. For instance, Bacco et al.~\cite{bacco_explainable_2021} used transformer-based extractive summarization of reviews, where the attention-selected sentences demonstrate how certain parts of the text contribute to the final output. This approach highlights the important segments of text and shows stakeholders why particular sentences were chosen. Similarly, the inclusion of attention-based or rationale-based explanations promotes transparency, enabling decision-makers to trust automated summaries in areas like market research or product feedback analysis.

Another core application is sentiment analysis, which is central to gauging public perception and refining marketing strategies. Although sentiment analysis can detect emotions or opinions in large-scale text data, it often operates as a ``black box.'' By applying explainability methods such as Local Interpretable Model-agnostic Explanations (LIME), Layer-wise Relevance Propagation (LRP), or attention-based heatmaps, organizations can better interpret why a particular sentiment label was assigned~\cite{capuano2021sentiment, du_techniques_2019, bacco_explainable_2021}. Surveys of interpretable machine learning catalogue these attribution families and the trade-offs between them~\cite{du_techniques_2019}.

Beyond sentiment analysis, XNLP also enhances customer support automation, including chatbots and self-service platforms that respond to consumer inquiries in real time. By making these systems explainable, companies can pinpoint why certain answers or suggestions were given, reducing user frustration and building trust~\cite{jenneboer2022impact, guerreiro2024systematic}. For example, the systematic literature review by Jenneboer et al.~\cite{jenneboer2022impact} found that the transparency and quality of chatbot responses strongly shape customer trust and loyalty. Table~\ref{tab:summary4} illustrates various CRM-related applications, outlining the models, explainability techniques, datasets, and performance metrics used.

\begin{table}[ht!]
\centering
\scriptsize
\renewcommand{\arraystretch}{0.9}
\caption{Summary of XNLP Applications in CRM}
\label{tab:summary4}
\setlength{\tabcolsep}{4pt}%
\begin{tabular}{P{0.55cm} P{1.3cm} P{1.1cm} P{1.3cm} P{1.0cm} P{1.1cm}}
\toprule
\textbf{Ref.} & \textbf{Application} & \textbf{Model} 
& \textbf{Explainability Method} & \textbf{Dataset} & \textbf{Metrics} \\ 
\midrule

\cite{capuano2021sentiment}
& Sentiment Analysis 
& BERT 
& LIME 
& Yelp Reviews 
& Accuracy, F1-score \\

\addlinespace
\cite{bacco_explainable_2021}
& Sentiment Analysis
& Cascaded Transformers
& Attention-based Extractive Summaries
& Movie Reviews
& Accuracy, F1-score, Precision \\

\addlinespace
\cite{jenneboer2022impact}
& Customer Support Automation
& Systematic Literature Review
& Analysis of Chatbot Transparency
& Published Chatbot Studies
& Customer Trust, Loyalty \\

\addlinespace
\cite{lei_rationalizing_2016}
& Multi-aspect Sentiment Analysis
& Generator--encoder (RNN)
& Extractive Rationales
& BeerAdvocate reviews
& Precision vs.\ human rationales \\

\addlinespace
\cite{yu2021understanding}
& Sentiment Analysis 
& CNN, BiGRU 
& Attention Heatmaps, Ablation 
& Amazon, BeerAdvocate 
& Accuracy, Precision, Recall, F1-score \\

\addlinespace
\cite{Antognini2021RationalizationTC}
& Multi-aspect Sentiment Analysis
& ConRAT (concept-based)
& Concept-level Rationales
& Beer Reviews, Amazon Electronics
& Accuracy, F1-score \\

\bottomrule
\end{tabular}
\end{table}

Empirical evaluations demonstrate the strong performance of transformer-based models in CRM applications. A fine-tuned BERT classifier reaches 92\% accuracy on aspect detection over Amazon product reviews, ahead of a T5 baseline at 91\%~\citep{alsulami2024sentiment}, while a hybrid transformer and BiLSTM model with an attention layer reports 94\% accuracy and 94\% macro F1 on tweet sentiment~\citep{li2024trabsa}. SHAP and LIME integration enables token-level attribution, revealing which words or phrases most influenced sentiment predictions, for instance identifying that adjectives describing product quality contribute more strongly to positive classifications than generic praise. This granular explainability proves particularly valuable in e-commerce settings, where understanding why a review was classified as negative enables targeted product improvements. The interpretable attention mechanisms in these hybrid architectures allow users to trace sentiment predictions back to specific textual features, though enhanced explainability introduces computational overhead during inference~\citep{li2024trabsa}.

\noindent
Overall, XNLP's integration into CRM marks a key shift in how businesses collect, interpret, and act on text data. By offering transparent rationales for automated summarization, sentiment detection, and response generation, these systems help organizations forge deeper connections with their customers. Explainability also does more than reassure users: it gives developers and stakeholders what they need to refine their NLP pipelines for accuracy and reliability. As shown in Table~\ref{tab:summary4}, a variety of architectures and techniques, ranging from BiLSTM to Transformer-based models, are being equipped with explainability features, enabling more interpretable and user-aligned CRM solutions. Persistent challenges remain: personalization pressures conflict with transparency, deployments increasingly span multiple languages, and customer-facing pipelines impose real-time constraints on explanation generation. LLM-based sentiment analysis likewise exhibits instability arising from prompt sensitivity and stochastic inference, and recent work positions explainability as a key mitigation for this variability~\citep{herrerapoyatos2025overview}.

\subsection{Chatbots and Conversational AI}
\label{sec:chatbots_conversational}

Chatbots and conversational AI systems have become increasingly prevalent in areas such as virtual assistance, customer support, and information retrieval. By using XNLP, these systems can offer more transparent and user-centric interactions, as their generated responses or recommendations can be coupled with clear rationales. This transparency builds user trust, enabling individuals to understand the why behind the bot's outputs and to feel more confident in adopting its suggestions or insights.

\noindent
%\textbf{Context-Aware Recommendations and Explainability.}
Context-aware recommendation systems aim to deliver personalized suggestions by incorporating elements of user intent, interaction history, or external data. Incorporating XNLP techniques into these recommenders can enhance user confidence by explicitly showing which contextual cues led to particular recommendations. For instance, the survey by Zhang et al.~\cite{zhang_deep_2019} reviews deep learning based recommender systems, including context-aware models whose attention mechanisms indicate how user-item interactions influence the outputs, which can support higher satisfaction.

\noindent
%\textbf{Advances in GPT-4 and GPT-4o.} 
In 2023, conversational AI took a major step forward with the introduction of GPT-4, known for its larger context window and improved accuracy in both English and code-based tasks. OpenAI's GPT-4o (``o'' for ``omni'') extended these advantages, matching GPT-4 Turbo performance in English while showing further gains in non-English languages~\cite{openai2024hello, openai2024mini}. Although these advanced models can generate more coherent and context-aware dialogue, they still require effective explainability strategies to clarify the internal reasoning paths or chain-of-thought that lead to each response. XNLP methods thus remain essential for building user trust, especially as chatbot complexity and capabilities continue to grow.

\noindent
%\textbf{Explainability and User Trust.} 
Recent studies stress the importance of explainable conversational AI for enhancing user satisfaction, transparency, and overall effectiveness~\cite{guerreiro2024systematic, soni2024impact, miglani-etal-2023-using}. While AI chatbots increasingly excel at generating swift, personalized responses, embedding XNLP capabilities can reveal how the system arrives at each answer. This can be accomplished through mechanisms like attention distributions, rationale highlighting, or feature visualization, each of which helps users grasp the underlying logic. Such interpretability also aids in detecting potential biases or inconsistencies, enabling developers to refine their models and ensuring the system behaves reliably under different conditions~\cite{cheng2022effects, Folstad2021chatbots}.

\noindent
Table~\ref{tab:summary5} highlights a range of applications in chatbots and conversational AI, detailing the models, explainability techniques, datasets, and evaluation metrics employed. These examples demonstrate how integrating transparency can bolster user trust, streamline recommendation processes, and refine user engagement strategies.

\begin{table}[ht!]
\centering
\scriptsize
\renewcommand{\arraystretch}{1}
\caption{Summary of XNLP Applications in Chatbots and Conversational AI}
\label{tab:summary5}
\setlength{\tabcolsep}{4pt}%
\begin{tabular}{P{0.55cm} P{1.3cm} P{1.1cm} P{1.3cm} P{1.0cm} P{1.1cm}}
\toprule
\textbf{Ref.} & \textbf{Application} & \textbf{Model} 
& \textbf{Explainability Method} & \textbf{Dataset} & \textbf{Metrics} \\ 
\midrule

\cite{li2025trust}
& Trust in Chatbots (Review)
& Various
& Synthesis of Trust Predictors
& 40 primary studies
& Trust antecedents, outcomes \\

\addlinespace
\cite{kovari2025explainable}
& Explainable Chatbots (Review)
& Knowledge graphs, LLMs
& Survey of Explanation Strategies
& N/A (Review)
& Trust, accountability \\

\addlinespace
\cite{zhang_deep_2019}
& Recommendation Systems (Survey)
& Deep Learning-based
& Attention Mechanism
& N/A (Survey)
& User Satisfaction \\

\addlinespace
\cite{guerreiro2024systematic}
& Chatbots in Business 
& Various 
& Transparency in Decision-Making
& N/A (Review)
& User Trust, Satisfaction \\

\addlinespace
\cite{ghazvininejad2018knowledge}
& Knowledge-Grounded Open-Domain Dialogue
& Seq2Seq
& External Knowledge Grounding
& Twitter, Foursquare
& BLEU, Diversity \\

\bottomrule
\end{tabular}
\end{table}

Empirical research on chatbot trust reveals that explainability significantly influences user acceptance and satisfaction. Systematic reviews synthesizing 40 studies identify five categories of trust predictors: user characteristics, machine attributes, interaction quality, social factors, and contextual elements~\citep{li2025trust}. Quantitative studies demonstrate that chatbot usability explains 59\% of the variance in user trust ($R^2 = 0.59$), which in turn shapes customer attitudes, behavioral intentions, and satisfaction~\citep{gu2023usability}. Across that literature, the quality of a system's explanations emerges as a positive predictor of trust and acceptance~\citep{li2025trust}. However, balancing explanation detail with conversational fluency remains challenging: overly verbose explanations can disrupt dialogue flow, while sparse explanations may fail to satisfy user curiosity about system behavior.

\noindent
Overall, the integration of XNLP into conversational AI has the potential to reshape user interactions by providing richer, more transparent exchanges. As large-scale models like GPT-4o continue to grow in complexity, designing dependable \emph{explainability} mechanisms will be central to maintaining user trust and improving dialogue outcomes. From clarifying context-aware recommendations to justifying chatbot responses, XNLP methods will shape the future trajectory of conversational systems. A recent review of explainable chatbots, spanning knowledge-graph hybrids and ChatGPT-style systems, identifies trust and accountability as the field's organizing concerns~\citep{kovari2025explainable}. Two challenges are particularly acute in this setting: explanations must be woven into dialogue without disrupting conversational flow, and rationales must be generated in real time alongside the response itself.

\subsection{Social and Behavioral Science}
\label{sec:social_behavioral_science}

Social and behavioral science research often addresses highly complex societal challenges, such as hate speech, sexism, misinformation, and emotional well-being. XNLP provides valuable tools for enhancing the reliability of automated methods in this domain, enabling clearer rationales behind model outputs. This transparency allows researchers and practitioners to gain deeper insights into underlying language patterns and the decision-making processes of NLP models. Table~\ref{tab:summary6} summarizes representative XNLP studies in this domain.

%\subsubsection{Hate Speech and Sexism Detection}
Transformer-based architectures, including BERT, have greatly improved the detection of offensive language. Mozafari et al.~\cite{mozafari2020hate} fine-tuned BERT for hate speech detection on Twitter data and added a bias-alleviation mechanism that reweights training samples, reducing the racial bias the classifier would otherwise inherit from its training set. Likewise, Saleh et al.~\cite{saleh2023detection} developed domain-specific word embeddings focused on hate speech data, offering insight into how such content manifests linguistically. Shared tasks have driven much of this work: the sEXism Identification in Social neTworks (EXIST) campaign covers English and Spanish, while SemEval-2023 Task 10 targets English content from Gab and Reddit with fine-grained, explainability-oriented labels~\cite{plaza2023overview, kirk2023semeval, mohammadi2023towards, mohammadi2024transparent}. These studies highlight the value of human-in-the-loop validation, where user feedback helps align model outputs with human judgments and mitigate biases~\cite{anjum2023hate}.

Benchmark datasets with integrated explainability annotations have proven valuable for advancing research in this area. The HateXplain dataset~\citep{mathew2021hatexplain} provides 20,000 social media posts annotated with three-way classification labels (hate, offensive, normal), target community identification, and human rationales indicating which text spans influenced labeling decisions. BERT-based models trained with these rationales achieve 0.698 accuracy and 0.687 F1-score, with AUROC of 0.851, modest improvements over standard BERT (0.690 accuracy, 0.674 F1). Critically, models incorporating human rationales during training demonstrate reduced unintended bias toward target communities. In the SemEval-2023 Task 10 competition on explainable sexism detection, the best systems reached 0.87 macro F1 on binary sexism detection, but only 0.73 on the four-way category task and below 0.57 on the fine-grained 11-class task~\citep{kirk2023semeval}. These results point to a key insight: models achieving high classification accuracy do not necessarily score well on explainability metrics such as plausibility (overlap with human rationales) and faithfulness (whether explanations reflect actual model reasoning). Extending this line of work beyond English, a recent multilingual hate-speech benchmark covering Hindi, Telugu, and English provides token-level human rationales and evaluates models on both plausibility and faithfulness~\citep{rehman2026xmutest}.

%\subsubsection{Fake News and AI-Generated Content Detection}
Misinformation poses an escalating threat in digital communication. XNLP offers techniques to identify fabricated or AI-generated text, thereby safeguarding the integrity of online information. Capuano et al.~\cite{capuano2023content} review the machine- and deep-learning approaches used for content-based fake news detection. Similarly, Mohammadi et al.~\cite{mohammadi_2023_10079010} created a multilingual ensemble model, tested under a shared task by Computational Linguistics in the Netherlands (CLIN), to discern AI-generated text from human-authored writing~\cite{fivez2024clin33}. Explainability cuts both ways in this setting: applying SHAP and LIME to a detector reveals which tokens drive its decision, and those same attributions can be used to rewrite the text so that it evades detection~\cite{mohammadi2025token}. Beyond purely text-based strategies, multimodal systems like SceneFND integrate textual, contextual, and visual cues, showing enhancements in identifying misinformation across diverse datasets~\cite{zhang2024scenefnd}.

%\subsubsection{Psychological Studies and Mental Health}
Though frequently classified under medicine (see \emph{\nameref{sec:medicine}}), research in social and behavioral contexts also applies XNLP to assess mental health conditions and public sentiment. Han et al.~\cite{han2023sensing} employed NLP features from social media posts to predict emotional or psychological well-being. Meanwhile, Ahmed et al.~\cite{AHMED2024} experimented with transformer-based architectures to classify depression severity, illustrating how explainable output can guide mental health interventions. Ali et al.~\cite{ali2022large} likewise analyzed sentiment at scale across tweets from the 2020 U.S. Presidential Elections, demonstrating how these techniques can capture public opinion dynamics.

%\subsubsection*{Misinformation and Fake-News Detection}

Wang et al.~\cite{wang2024explainable} introduce L-Defense, a framework that partitions crowd discourse on a news claim into two opposing camps. It then extracts salient evidence for each side and prompts a large language model (LLM) to debate and defend their respective narratives. The ``winning'' defense produces a concise natural-language explanation along with the final veracity label, achieving higher detection accuracy than previous methods and offering users a transparent, evidence-based rationale.

%\subsubsection*{Mental-Health Analysis and Clinical Decision Support}

Schmidl et al.~\cite{schmidl2024assessing} compare Claude 3 Opus with ChatGPT-4.0 on 50 consecutive primary head and neck cancer cases, scoring each model against a conventional multidisciplinary tumor board on its clinical recommendation, its explanation, and its summarization. Scoring the explanation as a first-class outcome alongside the recommendation is the point of interest here: a recommendation a clinician cannot interrogate is difficult to act on, however accurate it proves in aggregate. In parallel, Wang et al.~\cite{wang2024depression} study social-media language to evaluate depression severity, identifying the phrases that most influenced each assessment. Such research reflects an emerging focus on verifying the \emph{practical utility} of XNLP explanations, not merely their plausibility in real-world healthcare settings.

\noindent
\textbf{Domain-Specific Challenges.} The social and behavioral science domain presents challenges that distinguish it from other application areas~\citep{gongane2025decoding, rawat2024explainable}. Hate speech and offensive language manifest differently across cultures, languages, and platforms: what constitutes hate speech in one cultural context may be acceptable discourse in another, so XAI methods must account for these variations rather than applying universal standards~\citep{rawat2024lowresource}. Low-resource languages compound the difficulty because training data and lexicons are often unavailable. The models' own judgments carry cultural assumptions as well. A transparent chain-of-thought evaluation of 20 LLMs against World Values Survey responses found markedly closer alignment for Western regions (Pearson's $r = 0.82$ on average) than for non-Western regions ($r = 0.61$)~\citep{mohammadi2025evalmoraal}.

Annotation itself is a second source of difficulty. Ground-truth labels in hate speech datasets often reflect annotator demographics and perspectives. When models learn from biased annotations, their explanations may perpetuate those biases, and explaining why a model flagged content as hateful requires acknowledging the subjectivity of the labeling process. Perspectivist approaches that model annotator disagreement rather than aggregating it away are gaining traction, and such disagreement-aware systems call for dedicated interpretability methods~\citep{xu2026beyond}. How much of that disagreement is demographic is itself an empirical question: on a sexism-annotation corpus, annotator demographics accounted for only a small share of the variance in labels, with the content of the item dominating~\citep{mohammadi2025annotations}. Beyond annotation, users attempting to spread hate or misinformation continuously adapt their language to evade detection through coded language, deliberate misspellings, or memes. XAI methods must keep pace while explaining current patterns well enough that human moderators can recognize novel evasion tactics. Platform context matters too: the same text may read differently on Twitter, Reddit, or a messaging app, and explanations must incorporate thread structure, community norms, and user history to be actionable for moderators.

These challenges highlight that effective XNLP in social science requires not just technical sophistication but also deep engagement with the sociocultural contexts in which models are deployed.

\begin{table}[ht!]
\centering
\scriptsize
\renewcommand{\arraystretch}{1}
\caption{Summary of XNLP Applications in Social and Behavioral Science}
\label{tab:summary6}
\setlength{\tabcolsep}{4pt}%
\begin{tabular}{P{0.55cm} P{1.3cm} P{1.1cm} P{1.3cm} P{1.0cm} P{1.1cm}}
\toprule
\textbf{Ref.} & \textbf{Application} & \textbf{Model} 
& \textbf{Explainability Method} & \textbf{Dataset} & \textbf{Metrics} \\
\midrule

\cite{plaza2023overview}
& Sexism Detection 
& ML Algorithms 
& EXIST Competition 
& EXIST-2023 
& Classification Accuracy \\

\addlinespace
\cite{mozafari2020hate} 
& Hate Speech Detection 
& BERT + Bias Mitigation 
& Bias Mitigation Mechanism 
& Davidson \& Waseem 
& F1-measure \\

\addlinespace
\cite{saleh2023detection}
& Hate Speech Detection
& BERT \& BiLSTM
& Domain-Specific Word Embeddings
& Hate Speech Word Embedding
& Detection Accuracy \\

\addlinespace
\cite{kirk2023semeval}
& Explainable Sexism Detection (Shared Task)
& Participant systems
& Fine-grained Label Hierarchy
& EDOS (Gab, Reddit)
& Macro F1 \\

\addlinespace
\cite{mohammadi2024transparent}
& Explainable Sexism Detection 
& Ensemble (BERT, XLM-R, DistilBERT) 
& SHAP 
& EXIST-2023 
& Token Influence Analysis \\

\addlinespace
\cite{mohammadi_2023_10079010}
& AI-Generated Text Detection 
& Ensemble + Multilingual BERT 
& SHAP 
& Various Genres 
& Detection Accuracy \\

\addlinespace
\cite{han2023sensing}
& Mental Health Analysis 
& NLP Techniques 
& Feature Importance 
& Social Media Posts 
& Predictive Power \\

\addlinespace
\cite{wang2024explainable}
& Fake-News Detection (L-Defense)
& LLM-based Debate
& Natural-Language Justification
& Crowd Discourse
& Detection Accuracy \\

\addlinespace
\cite{rana2022rerrfact}
& Scientific Claim Verification
& RoBERTa (Modular Pipeline)
& Rationale Selection, Stance Prediction
& SciFact
& F1-score \\

\addlinespace
\cite{schmidl2024assessing}
& Tumor-Board Treatment Plans 
& GPT-4, Claude 3 
& Step-by-step Rationales 
& Clinical Case Reports 
& Expert-Level Validity \\

\addlinespace
\cite{wang2024depression}
& Depression Severity Assessment 
& Transformer-based 
& Highlighted Key Phrases 
& Social Media Posts 
& Clinical Utility \\

\bottomrule
\end{tabular}
\end{table}

\noindent
Overall, integrating XNLP into social and behavioral science marks a major advance in addressing issues such as hate speech, sexism, misinformation, and mental health monitoring~\cite{mathew2021hatexplain, yang2023hareexplainablehatespeech}. By using the interpretability features of advanced transformer models, domain-specific embeddings, and multilingual datasets, researchers can develop classifiers that identify problematic content or behavior and explain why they flagged it. This approach reinforces the credibility of automated systems and aids stakeholders, ranging from social scientists to clinicians, in better understanding the subtle language patterns that drive complex social phenomena.

\subsection{Human Resources (HR)}
\label{sec:hr}

HR covers processes ranging from talent acquisition and employee sentiment analysis to performance reviews and diversity initiatives. XNLP has recently gained traction in optimizing these functions, offering transparent decision-making tools that can build trust, reduce biases, and streamline various HR operations.

%\subsubsection{Talent Acquisition and Recruitment}
Transformer-based architectures like BERT and GPT are increasingly used to automate the resume-screening and candidate-ranking process, matching required skills more effectively with job descriptions~\cite{patwardhan2023transformers}. For instance, Gan et al.~\cite{gan2024application} demonstrate how an LLM-agent framework that summarizes and grades each resume can make the initial candidate-to-job matching phase more efficient. In parallel, the survey by Gurrapu et al.~\cite{gurrapu2023rationalization} shows how rationalization techniques can explain model outputs, a capability that can enhance stakeholder trust in the screening process. Akkasi~\cite{AKKASI2024} further highlights how a transformer-based ensemble model can accurately extract both technical and non-technical competencies, improving the precision and transparency of candidate matching.

%\subsubsection{Employee Sentiment Analysis}
In understanding workforce morale, XNLP provides sentiment analysis techniques to analyze feedback from surveys, internal forums, and social media. As the state-of-the-art review by Jim et al.~\cite{jim2024recent} shows, transformer-based models like RoBERTa can process large volumes of comments, capabilities that HR teams can use to pinpoint negative sentiments and emerging workplace issues. Attention-driven methods also clarify which textual factors influenced each sentiment label, allowing HR departments to better understand employee concerns~\cite{danilevsky2020survey}.

%\subsubsection{Performance Evaluation}
Traditional performance reviews often lack consistency and can harbor biases. By incorporating XNLP to analyze written feedback and performance data, organizations achieve more objective and transparent evaluations. As recent overviews of NLP developments note~\cite{chang2023natural}, attention mechanisms in transformer models can highlight the most relevant textual feedback, offering a potential rationale for performance scores. Work on human-annotated rationales for text classification~\cite{herrewijnen2023human} similarly suggests that grounding model explanations in human reasoning can make automated assessments easier for HR managers to interpret.

%\subsubsection{Diversity and Inclusion}
XNLP also contributes to diversity and inclusion within companies by detecting linguistic biases in job postings, internal policies, and employee communications. Audits of commercial screening systems show why this matters. Bias surfaces in how the underlying model ranks candidates, not in anything the system reports about itself. Interpretable outputs are what make the problem visible to an organization at all~\cite{kaushal2024resume, webster2025fairness}. This transparency helps organizations address the root causes of inequality and maintain more inclusive hiring practices.

Bias in AI-driven recruitment remains a substantial problem, however. Large-scale evaluations reveal gender, racial, and intersectional bias in AI resume screening. In an audit of text-embedding retrieval models over more than 500 resumes, White-associated names were favored in 85.1\% of comparisons and female-associated names in only 11.1\%. Black male candidates were disadvantaged in up to 100\% of the intersectional comparisons tested~\citep{kaushal2024resume}. The ``black-box'' nature of such systems complicates bias auditing, because disparities of this kind surface only under systematic probing and are not visible in anything the system reports about its own decision. Regulatory frameworks are emerging in response: New York City mandates auditing of AI hiring systems since 2023, while Colorado's comprehensive AI employment law takes effect in 2026. These developments point to the need for XNLP methods that improve recruitment efficiency while still producing transparent, auditable decision rationales that satisfy ethical standards and legal requirements. A 2025 audit of eight commercial resume-screening platforms reinforces this need, proposing dual validation of both bias and competence for such systems~\citep{webster2025fairness}. Recent interpretability work adds a further warning: anti-bias prompting can fail on realistic resumes, while edits at the activation level cut demographic bias substantially. The bias never shows up in the model's stated reasoning, which connects directly to the faithfulness concerns raised in \emph{\nameref{subsec:cot_faithfulness}}~\citep{karvonen2025robustly}.

\begin{table}[ht!]
\centering
\scriptsize
\renewcommand{\arraystretch}{0.9}
\caption{Summary of XNLP Applications in HR}
\label{tab:summaryHR}
\setlength{\tabcolsep}{4pt}%
\begin{tabular}{P{0.55cm} P{1.3cm} P{1.1cm} P{1.3cm} P{1.0cm} P{1.1cm}}
\toprule
\textbf{Ref.} & \textbf{Application} & \textbf{Model} 
& \textbf{Explainability Method} & \textbf{Dataset} & \textbf{Metrics} \\ 
\midrule

\cite{gan2024application}
& Resume--Job Matching
& LLM Agents
& Generated Summaries and Grades
& Resume Data
& Match Accuracy \\

\addlinespace
\cite{kaushal2024resume}
& Recruitment Bias Auditing
& Text-Embedding Retrieval
& Bias Audits
& Resumes, Job Descriptions
& Selection-Rate Disparity \\

\addlinespace
\cite{AKKASI2024}
& Skill Extraction 
& Transformer Ensemble 
& Attention Mechanisms 
& Job Descriptions 
& Precision, Recall \\

\addlinespace
\cite{jim2024recent}
& Sentiment Analysis (Review)
& Various
& Survey of Methods
& N/A (Review)
& Sentiment Insights \\

\addlinespace
\cite{herrewijnen2023human}
& Explainable Text Classification (Survey)
& Various
& Human-Annotated Rationales
& N/A (Survey)
& Plausibility, Faithfulness \\

\bottomrule
\end{tabular}
\end{table}
\noindent
As illustrated in Table~\ref{tab:summaryHR}, the use of XNLP across various HR functions, from screening resumes to promoting inclusivity, shows its capacity to bring objectivity, transparency, and efficiency to people-centric tasks. Studies consistently show that explainability features such as attention heatmaps and rationale generation strengthen user trust and align HR strategies more closely with organizational values. Looking ahead, ongoing research into transformer-based and explainable methods promises further refinements in reducing biases, understanding employee concerns, and improving overall talent management.

\subsection{Other Applications}
\label{sec:other_applications}

Beyond the domains discussed earlier, XNLP finds utility in a wide range of tasks and sectors. These include language generation, text classification, machine translation, summarization, visual question answering, and more. In such contexts, explainability is what makes it possible to see how and why a model produces specific outputs, offering transparency and trustworthiness in settings that may require critical decision-making or that deal with large-scale user interactions.

%\paragraph{Language Generation and Bias Detection.}

Sheng et al.~\cite{sheng2019woman} prompted language models such as GPT-2 with mentions of different demographic groups and introduced the notion of \emph{regard} towards a demographic as a bias metric, training a classifier to score generated text at scale. Their analysis shows how societal biases surface in generation, and how far sentiment serves as a proxy for the harm actually expressed.

%\paragraph{Rationale-Based Explanations.}

DeYoung et al.~\cite{deyoung2019eraser} proposed rationale-based explanations for text classification tasks across multiple datasets (BoolQ, e-SNLI, etc.), evaluating the quality of explanations on fidelity, comprehensiveness, and sufficiency. Similarly, Jacovi and Goldberg~\cite{jacovi2020towards} contributed a position paper on how the faithfulness of explanations should be defined and evaluated, while Ayyubi et al.~\cite{ayyubi2020generating} extended rationale generation to visual question answering with multimodal transformer models.

%\paragraph{Transformer Analyses and Multimodal Tasks.}
Several works have examined annotation of word importance and visualization of model internals. For instance, He et al.~\cite{he2019towards} annotated tokens in neural machine translation with gradient-based importance weights, testing Transformer and RNN systems on WMT benchmarks. Alishahi et al.~\cite{alishahi2019analyzing} surveyed such analysis methods for neural language models in their overview of the first BlackboxNLP workshop. Likewise, Hoover et al.~\cite{hoover-etal-2020-exbert} employed BERT and other Transformer variants to study learned self-attention, providing interactive visual analyses of attention patterns, and Zhou et al.~\cite{zhou2020towards} generated natural-language explanations as latent variables to interpret model decisions. Beyond purely text-based challenges, multimodal Transformer approaches such as those of Ayyubi et al.~\cite{ayyubi2020generating} and Tang et al.~\cite{tang2021cognitive} addressed visual commonsense reasoning, indicating how explainable components can strengthen cross-domain understanding.

%\paragraph{Commonsense Reasoning and Quality Estimation.}
Commonsense-related tasks and interpretability-oriented methods have also drawn increasing attention. For example, Lakhotia et al.~\cite{lakhotia2020fid} introduced FiD-Ex to generate extractive rationales, while Wiegreffe et al.~\cite{wiegreffe2020measuring} used T5-based joint models to produce free-text explanations in tasks like commonsense question-answering and natural language inference. Additional works by Plyler et al.~\cite{plyler2021making},~\cite{fomicheva-etal-2022-translation}, and~\cite{chan2022unirex} concentrated on counterfactually augmented rationale-based text classification, translation quality estimation, and universal rationalization frameworks, respectively. Table~\ref{tab:summary7} highlights diverse XNLP applications along with key models, methods, datasets, and metrics.

\begin{table}[ht!]
\centering
\scriptsize
\renewcommand{\arraystretch}{0.9}
\caption{Summary of XNLP Studies in Other Applications}
\label{tab:summary7}
\setlength{\tabcolsep}{4pt}%
\begin{tabular}{P{0.55cm} P{1.3cm} P{1.1cm} P{1.3cm} P{1.1cm} P{1.1cm}}
\toprule
\textbf{Ref.} & \textbf{Application} & \textbf{Model} 
& \textbf{Explainability Method} & \textbf{Dataset} & \textbf{Metrics} \\ 
\midrule

\cite{sheng2019woman}
& Language Generation
& GPT-2, etc.
& Regard Classifier, Sentiment Scoring
& Prompted LM generations
& Regard, sentiment scores \\

\addlinespace
\cite{he2019towards}
& Neural Machine Translation
& Transformer, RNN
& Gradient-based Word Importance
& WMT Zh--En, En--Fr, En--Ja
& BLEU under Perturbation \\

\addlinespace
\cite{deyoung2019eraser}
& Text Classification \& Rationale 
& Various Models 
& Rationale-based Explanations 
& BoolQ, e-SNLI, etc. 
& Fidelity, Comprehensiveness \\

\addlinespace
\cite{jacovi2020towards}
& Faithfulness Evaluation (Position Paper)
& N/A
& Criteria for Faithful Explanations
& N/A
& N/A \\

\addlinespace
\cite{ayyubi2020generating}
& Visual Question Answering 
& Multimodal Transformer 
& Rationale Generation 
& VQA v2.0, VizWiz 
& Answer Accuracy \\

\addlinespace
\cite{alishahi2019analyzing}
& Analysis of Neural Language Models (Workshop Overview)
& Various
& Survey of Analysis Methods
& N/A
& N/A \\

\addlinespace
\cite{zhou2020towards}
& Natural Language Understanding
& Pre-trained LMs
& Natural-Language Explanations as Latent Variables
& --
& -- \\

\addlinespace
\cite{hoover-etal-2020-exbert}
& Transformer Model Analysis
& BERT, etc.
& Visualization, Self-Attention
& --
& -- \\

\addlinespace
\cite{jain_attention_2019}
& Attention Analysis (Classification, QA)
& BiLSTM, CNN
& Attention Distributions
& SST, IMDb, MIMIC
& Kendall's Tau Correlation \\

\addlinespace
\cite{lakhotia2020fid}
& Extractive Rationale Generation
& Seq2seq (Fusion-in-Decoder)
& FiD-Ex Sentence Markers
& ERASER (QA, fact verification)
& Task F1, Rationale F1 \\

\addlinespace
\cite{tang2021cognitive}
& Visual Commonsense Reasoning 
& DMVCR 
& Dynamic Working Memory 
& VCR, RevisitedVQA
& -- \\

\addlinespace
\cite{wiegreffe2020measuring}
& Free-text Rationales 
& T5-based Joint Models 
& Natural Language Rationales 
& Commonsense QA, NLI 
& Feature Importance Agreement \\

\addlinespace
\cite{plyler2021making}
& Rationale-based Text Classification
& Rationale Extraction Models
& Counterfactual Data Augmentation
& Beer, Hotel Reviews
& -- \\

\addlinespace
\cite{fomicheva-etal-2022-translation}
& Translation Error Detection
& Multilingual QE models
& Feature Attribution as Rationales
& MLQE-PE
& Pearson, Kendall's Tau \\

\addlinespace
\cite{chan2022unirex}
& Text Classification 
& UniREX 
& Rationalization 
& FEVER, Movie Reviews 
& Precision, Recall, F1 \\

\bottomrule
\end{tabular}
\end{table}

\noindent
%\textbf{Concluding Observations.} 
XNLP's application in these diverse areas reflects its growing impact, including tasks that may not carry as high a risk as medical or financial domains but still benefit from transparency (e.g., language bias detection, rationale generation, and VQA). By illuminating \emph{how} decisions are reached, XNLP makes NLP-driven solutions easier to trust, particularly in critical tasks such as summarization, machine translation, or text classification, where hidden biases could otherwise remain undiscovered. The consistent theme across studies is that explainability strengthens user confidence and supports auditing and refining model performance.

\noindent
%\textbf{Comparisons Across Domains.} 
From the compiled studies, we observe that \emph{risk-sensitive} areas (e.g., healthcare) rely strongly on actionable explanations (feature importance, rule extraction) to ensure trustworthy decision-making. In contrast, sectors like finance emphasize attention-based methods for spotting anomalies in vast data streams. CRM and chatbot applications often focus on generating user-facing explanations, thereby building trust in real-time interactions. Regardless of domain, the shared principle is that \emph{explainability} underpins system reliability and user acceptance, paving the way for broader adoption of NLP-based automation.

\subsection{Cross-Domain Analysis of Divergent Explainability Requirements}
\label{subsec:cross_domain}

While the preceding sections examined XNLP applications within individual domains, a cross-domain perspective reveals both common threads and critical differences in how explainability is conceptualized, implemented, and evaluated. This synthesis addresses a key gap in the literature: the lack of systematic comparison across domains~\citep{amparore2024comprehensive, zhang2023m4}. Table~\ref{tab:cross_domain} provides a comparative overview of domain-specific XNLP requirements.

\begin{table*}[ht!]
\centering
\scriptsize
\renewcommand{\arraystretch}{1.1}
\caption{Cross-Domain Comparison of XNLP Requirements, Methods, and Challenges}
\label{tab:cross_domain}
\begin{tabular}{p{1.8cm} p{2.2cm} p{2.5cm} p{2.2cm} p{4.5cm}}
\toprule
\textbf{Domain} & \textbf{Primary XAI Need} & \textbf{Preferred Methods} & \textbf{Key Metrics} & \textbf{Unique Challenges} \\
\midrule

\textbf{Medicine}
& Clinical actionability; patient safety
& Feature importance, rule extraction, attention visualization
& AUC, clinical validity, fidelity, user trust
& HIPAA/GDPR privacy constraints; concept drift in EHR data; class imbalance; integration with clinical workflows~\citep{nazir2025xai, noor2025unveiling} \\
\addlinespace

\textbf{Finance}
& Regulatory compliance; fraud detection
& SHAP, LIME, attention heatmaps, ensemble methods
& AUC-ROC, accuracy, compliance score
& Adversarial gaming of explanations; real-time scalability; temporal dynamics in markets~\citep{vcernevivciene2024explainable, hassan2025fraud} \\
\addlinespace

\textbf{Systematic Reviews}
& Reproducibility; efficiency
& Rule-based extraction, feature highlighting
& WSS, inter-rater agreement ($\kappa$), recall
& Scalability across heterogeneous corpora; integration with existing review tools~\citep{ferdinands2023active, teijema2025simulation} \\
\addlinespace

\textbf{CRM}
& User satisfaction; personalization
& Attention visualization, rationale generation, dialogue explanations
& Accuracy, F1-score, satisfaction scores
& Personalization vs.\ transparency trade-off; multi-language requirements; real-time response needs~\citep{herrerapoyatos2025overview} \\
\addlinespace

\textbf{Chatbots}
& User understanding; trust building
& Dialogue-level explanations, attention visualization
& User trust, satisfaction, task completion rate
& Balancing explanation detail with conversational flow; real-time generation~\citep{gu2023usability} \\
\addlinespace

\textbf{Social \& Behavioral Science}
& Content moderation; safety
& SHAP, LIME, attention-based methods
& Accuracy, F1-score, fairness metrics
& Cultural sensitivity in hate speech detection; annotation bias; platform-specific language~\citep{gongane2025decoding, rawat2024explainable} \\
\addlinespace

\textbf{HR}
& Fair hiring; bias reduction
& Feature importance, counterfactual explanations
& Match accuracy, bias metrics, fairness scores
& Legal liability for biased decisions; sensitive personal data; cross-cultural validity of criteria~\citep{webster2025fairness} \\

\bottomrule
\end{tabular}
\end{table*}

\noindent
\textbf{Convergent Themes.} Despite domain-specific variations, several themes emerge across all application areas. Trust is the universal goal: across medicine, finance, HR, and social science, building user confidence in model predictions is the shared aim, although what counts as trustworthy varies. Clinicians require explanations that map to medical concepts~\citep{abdullah2025chatgpt}, while financial regulators demand audit trails that satisfy compliance frameworks~\citep{vcernevivciene2024explainable}. Feature-importance methods dominate: SHAP and LIME remain the most widely adopted methods across domains because they are model-agnostic and their outputs are easy to read, but their computational cost limits real-time deployment in finance and CRM settings~\citep{hassan2025model}. Finally, an evaluation gap recurs: technical metrics such as fidelity and faithfulness remain disconnected from practical utility. Unified benchmarks like M$^4$~\citep{zhang2023m4} and the fidelity meta-evaluation of Amparore et al.~\cite{amparore2024comprehensive} address this gap, but domain-specific validation remains essential (see \emph{\nameref{sec:critical}} and Table~\ref{tab:domain_eval}).

\noindent
\textbf{Divergent Requirements.} The analysis also reveals fundamental differences. Risk tolerance varies: medicine and finance operate under strict regulatory oversight with little room for unexplained decisions, while CRM applications prioritize user experience over exhaustive explanations. Temporal dynamics differ as well: financial applications face rapidly changing data distributions and adversarial actors who try to game explanations, whereas medical applications see slower concept drift but stricter validation requirements. Stakeholders also differ: healthcare explanations must serve clinicians, patients, and regulators at once, each with different needs and expertise, while HR systems mainly serve recruiters and compliance officers. And the useful granularity of an explanation is domain-specific: social science applications such as hate speech and misinformation detection often need token-level explanations of why particular phrases triggered a classification, while systematic reviews benefit from document-level summaries.

\noindent
\textbf{Methodological Implications.} These cross-domain insights suggest that a ``one-size-fits-all'' approach to XNLP is insufficient. Future research should develop domain-adaptive explanation frameworks that can adjust their granularity, format, and validation criteria based on the specific deployment context. The growing availability of domain-specific benchmarks and evaluation frameworks~\citep{zhang2023m4, amparore2024comprehensive} provides a foundation for such adaptive approaches. Recent consolidation efforts that unify dozens of XAI metric groups into a single evaluation framework~\citep{dembinsky2026unifying}, and expert-interview studies mapping EU AI Act interpretability mandates to concrete practice~\citep{dhaini2025from}, reinforce the case for such domain-aware standardization.

\noindent
\textbf{Where improvements are most needed.} The cross-domain comparison also exposes uneven methodological maturity: HR and CRM studies rarely evaluate the faithfulness of the explanations they deploy; medicine lacks standardized protocols for establishing clinical validity of explanations; finance requires explanations that remain robust under adversarial gaming; and systematic-review automation still lacks transparent, auditable stopping criteria~\citep{kempny2026when}.

The empirical evidence presented throughout these application domains reveals that explainability contributes measurably to system effectiveness. In healthcare, RETAIN reaches an AUC of 0.8705 for heart failure prediction, matching an unconstrained RNN baseline, while keeping the attention mechanism clinicians would audit~\citep{choi_retain_2016}. Financial fraud detection systems combining ensemble methods with SHAP explanations report high accuracy while maintaining interpretability, demonstrating that transparency need not compromise performance~\citep{hassan2025fraud}. In CRM contexts, BERT-based aspect detection on product reviews reaches 92\% accuracy~\citep{alsulami2024sentiment}, and an attention-enhanced hybrid transformer reports 94\% macro F1 on tweet sentiment~\citep{li2024trabsa}. Active learning approaches in systematic reviews substantially shorten the time to discover relevant records, while bias assessment tools demonstrate inter-rater reliability ($\kappa$ from 0.42 to 0.48) comparable to human reviewers~\citep{ferdinands2023active, marshall_robotreviewer_2016}. Social science applications benefit from human rationales, with models trained on HateXplain achieving 0.698 accuracy and 0.687 F1-score~\citep{mathew2021hatexplain}. Chatbot research demonstrates that usability explains 59\% of variance in user trust ($R^2 = 0.59$), highlighting the connection between explainability and user acceptance~\citep{gu2023usability}. In HR, documented biases in AI resume screening, where embedding-based retrieval favored White-associated names in 85.1\% of comparisons, make explainability essential for bias auditing and for fair hiring outcomes~\citep{kaushal2024resume}. Collectively, these findings establish that explainability serves not merely as a desirable feature but as a measurable contributor to the reliability, fairness, and effectiveness of NLP systems in practice. These figures, however, originate from heterogeneous datasets and evaluation protocols and are mostly task-performance rather than explanation-quality metrics; they indicate feasibility, not comparability.

\section{Critical Aspects of XNLP}
\phantomsection
\label{sec:critical}

\subsection{Evaluation Metrics for Quantifying Understanding}
\label{sec:evaluation}

What does it mean to \emph{understand} a model's output? In order to evaluate the effectiveness of XNLP techniques, it is necessary to \emph{quantify} the level of comprehension provided by the model's explanations, typically through a blend of quantitative and qualitative metrics~\cite{fan_interpretability_2020}. While many of these metrics can be described conceptually, formalizing them mathematically can offer a more precise view of how such measures are actually computed.

\subsubsection*{Quantitative Metrics}

These approaches aim to measure how closely an explanation mirrors a model's underlying reasoning, often focusing on the following concepts:

%\paragraph{Fidelity.} 
\textbf{Fidelity} reflects how accurately an explanation captures the \emph{true} behavior of a model. High-fidelity explanations yield the same predictions as the original model across an evaluation set. Formally, suppose \(\mathcal{D} = \{(x_i, y_i)\}_{i=1}^n\) is a dataset of \(n\) instances, where \(x_i\) is an input and \(y_i\) is the model's predicted label. Let \(M\) be the black-box model and \(E\) be the explainability function (e.g., a local surrogate or rationale generator). One way to measure fidelity is to track how well the surrogate's output \(\hat{y}_i = E(x_i)\) aligns with \(y_i = M(x_i)\). A simplified discrete version might be:

{\footnotesize
\begin{equation}\label{eq:fidelity}
  \text{Fidelity}(E, M, \mathcal{D}) \;=\; \frac{1}{|\mathcal{D}|}\sum_{(x_i,y_i)\in \mathcal{D}} \mathbf{1}\bigl[\hat{y}_i = y_i\bigr],
\end{equation}
}

where \(\mathbf{1}[\cdot]\) is the indicator function. The higher this value, the more the explanation's outcomes match the original model's predictions. For instance, LIME~\cite{ribeiro_why_2016} can approximate \(M\) locally with a simpler model to gauge this agreement.

%\paragraph{Coherence.} 
\textbf{Coherence} assesses the logical consistency or readability of an explanation in natural language form. Established language metrics (e.g., BLEU~\cite{papineni_bleu_2002}, ROUGE~\cite{lin_rouge_2004}, BERTScore~\cite{kaster-etal-2021-global}) are commonly used. For instance, the BLEU score between an automatically generated explanation \(E(x)\) and a reference explanation \(R(x)\) can be expressed as:

{\footnotesize
\begin{equation}\label{eq:bleu}
  \text{BLEU} \;=\; \exp \Bigl(\min\bigl(0,1-\tfrac{r}{c}\bigr) + \sum_{n=1}^N w_n \ln p_n \Bigr),
\end{equation}
}
where \(r\) is the reference length, \(c\) is the candidate length, \(p_n\) is the precision of matched n-grams, and \(w_n\) are weights (often uniform). Higher BLEU scores indicate closer alignment between generated explanations and reference texts. Importantly, coherence captures plausibility rather than faithfulness: a fluent, coherent rationale can misrepresent the model's actual computation (see \emph{\nameref{subsec:cot_faithfulness}}).

%\paragraph{Completeness.}
\textbf{Completeness} determines whether an explanation includes all salient factors behind a decision. One example arises in SHAP~\cite{lundberg2017unified}, where the contributions \(\phi_i\) of a feature set \(N\) together account for the gap between the model's expected output and its prediction \(M(x)\) on this input. A simplified Shapley-based formula is:

{\footnotesize
\begin{equation}\label{eq:shap}
  \phi_i(M, x) \;=\; \sum_{S \subseteq N\setminus \{i\}} \frac{|S|!\,(|N|-|S|-1)!}{|N|!}\Bigl(M(S \cup \{i\}) - M(S)\Bigr).
\end{equation}
}

The efficiency axiom then gives \(M(x) = \phi_0 + \sum_{i=1}^{|N|} \phi_i\), where \(\phi_0 = \mathbb{E}[M(x)]\) is the model's expected output over the background data. Completeness is the requirement that the attributions leave none of that gap unassigned. Extending this axiomatic treatment from individual features to \emph{interactions} between them requires additional assumptions, and the choice of those assumptions determines which interaction index one obtains~\cite{tsai2023faith}. Explanations lacking key contributions may have lower completeness scores.

\subsubsection*{Qualitative Metrics}

While quantitative metrics address how well the explanation aligns with the model's internal logic or reference texts, qualitative measures focus on user perceptions and real-world usability:

%\paragraph{User Trust.}
\textbf{User trust}~\cite{lipton2018mythos} gauges how confident users feel about an explanation. It can be approached via questionnaires or controlled user tests, but is not usually reduced to a single formula. Instead, a rating scale (e.g., 1 to 5) is often used to approximate trust. \textbf{Satisfaction} deals with how helpful or intuitive users find an explanation~\cite{doshi2017towards}. It may be measured as a difference in user performance or preference when they have access to explanations versus when they do not. \textbf{Transparency} captures whether users believe they understand the model's reasoning. This can be elicited via self-report items: for instance, ``On a scale of 1 to 7, rate how well you grasp the model's decision rationale.''~\cite{lipton2018mythos}. Aggregated responses create a transparency index, though it remains subjective by nature. Table~\ref{tab:evaluation_metrics} consolidates these quantitative and qualitative metrics.

\begin{table*}[ht!]
\centering
\scriptsize
\renewcommand{\arraystretch}{0.9}
\caption{Overview of Evaluation Metrics for XNLP Techniques}
\label{tab:evaluation_metrics}
\begin{tabular}{p{2cm} p{2cm} p{6.5cm} p{1.5cm}}
\toprule
\textbf{Type} & \textbf{Metric} & \textbf{Description} & \textbf{Study} \\
\midrule

\multirow{3}{*}{\textbf{Quantitative}} 
 & Fidelity 
 & Measures how accurately the explanations reflect the model's behavior; see Eq.~\ref{eq:fidelity} for a simplified version. 
 & \citep{ribeiro_why_2016} \\
 & Coherence
 & Assesses clarity and logical consistency in generated explanations (e.g., BLEU in Eq.~\ref{eq:bleu}).
 & \citep{papineni_bleu_2002} \\
 & Completeness
 & Evaluates whether all relevant factors are included, e.g., Shapley-based sums in Eq.~\ref{eq:shap}. 
 & \citep{lundberg2017unified} \\

\midrule

\multirow{3}{*}{\textbf{Qualitative}}
 & User Trust
 & Measures the confidence users have in the model's explanations (often via surveys). 
 & \citep{lipton2018mythos} \\
 & Satisfaction
 & Gauges user acceptance and perceived usefulness; frequently collected through rating scales.
 & \citep{doshi2017towards} \\
 & Transparency
 & Evaluates whether users feel they understand the model's reasoning.
 & \citep{lipton2018mythos} \\

\bottomrule
\end{tabular}
\end{table*}

In practice, both quantitative and qualitative assessments are needed to form a full picture of an explanation's quality~\cite{hoffman2018metrics, mohseni2021multidisciplinary}. Some work goes further and folds human judgment directly into model selection, optimizing for explanations that people can actually use rather than for a proxy metric~\cite{lage2018human}. However, the reliance on subjective judgments can introduce bias, and users may feel satisfied with explanations that \emph{appear} coherent but are not causally accurate~\cite{miller2019explanation}. This tension is part of what makes evaluating XNLP methods in real-world settings difficult. S{\o}gaard~\cite{sogaard2021explainable} emphasizes that context-aware metrics, those tailored to specific domain needs, may further improve the relevance and robustness of XNLP evaluations.

\noindent
Building XNLP systems often entails balancing model performance with interpretability. Although simpler models (e.g., linear regression, decision trees) are more transparent, they may fail to capture the complexities of language data as accurately as neural networks~\cite{doshi2017towards}. In high-stakes domains like healthcare and finance, navigating this \emph{explainability-performance} trade-off is critical; inaccurate predictions can have serious real-world consequences, but opaque models are less likely to be trusted by stakeholders. 

\noindent
\textbf{Time Complexity.} Some XNLP methods, notably perturbation- and sampling-based approaches such as SHAP (Eq.~\ref{eq:shap}), require repeated model evaluations and can incur overhead comparable to the training phase itself. Others, like attention-based mechanisms, exploit existing model components to highlight important tokens with less added cost. As model sizes grow~\cite{brown_language_2020}, the cost of producing explanations grows with them, and perturbation-based methods that require many forward passes per instance scale poorly~\citep{hassan2025model}. Scalability thus becomes a key bottleneck for widespread adoption of XNLP, particularly when handling large datasets or complex tasks.

\noindent
\textbf{Scalability and Standardization.} Large-scale LLMs have heightened concerns about the feasibility of real-time explanations~\cite{hassan2025model}. Ongoing research focuses on optimizing explanation modules, exploring ways to produce faithful insights without unduly burdening memory or compute cycles~\cite{sogaard2021explainable}. Meanwhile, standardized evaluation protocols are needed to compare XNLP approaches across domains, ensuring consistency in how metrics (e.g., fidelity, completeness) are applied.

\subsubsection*{Fidelity, Faithfulness, and Standardized Benchmarks}

A fundamental conceptual issue in XNLP evaluation is the distinction between \emph{fidelity} and \emph{faithfulness}~\citep{amparore2024comprehensive}. While often used interchangeably, these concepts address different aspects of explanation quality. Fidelity measures whether an explanation accurately predicts the model's outputs: whether a surrogate model or attribution method can reproduce the original model's behavior on test inputs. Faithfulness asks whether the explanation reflects the actual causal mechanisms the model uses to arrive at its predictions, not merely correlations or post-hoc rationalizations.

This distinction matters greatly. An explanation can have high fidelity (accurately predicting model outputs) while having low faithfulness (not representing the true reasoning process). The classic demonstration is the set of sanity checks of~\cite{adebayo2018sanity}. Several widely used saliency methods produce convincing-looking maps that barely change when the model's weights or the training labels are randomized. A map that survives randomization cannot be reporting what the trained model computes. Recent work on Chain-of-Thought faithfulness~\citep{turpin2023language, lanham2023measuring} demonstrates this gap: LLMs can generate plausible reasoning chains that correctly predict outputs but do not reflect their internal computations. Similarly, attention weights may correlate with model predictions without causally determining them~\citep{jain_attention_2019}. Faithfulness measurement is itself maturing: causal-counterfactual metrics can quantify the faithfulness of self-explanations and detect rationales that conceal the influence of social biases~\citep{matton2025walk}, while model-editing-based meta-evaluation assesses which faithfulness metrics reliably discriminate faithful from unfaithful explanations~\citep{zaman2025causal}.

A persistent challenge in applying such tests is the out-of-distribution (OOD) perturbation problem~\citep{xu2024ffidelity}. Traditional fidelity metrics perturb input features to measure their importance, but these perturbations often create inputs that fall outside the model's training distribution, leading to unreliable assessments. Zheng et al.~\cite{xu2024ffidelity} propose F-Fidelity, a framework that explicitly accounts for distribution shifts during evaluation, providing more reliable fidelity estimates for text classifiers.

The lack of standardized evaluation has long hindered progress in XNLP, and unified benchmarks have begun to close this gap. M$^4$~\citep{zhang2023m4} evaluates explanation methods across text, image, and multimodal domains under standardized faithfulness metrics, and finds that many popular methods perform inconsistently across modalities, which argues for domain-adaptive evaluation. Amparore et al.~\cite{amparore2024comprehensive} instead benchmark the fidelity metrics themselves rather than the explanation methods; this meta-evaluation finds substantial disagreement between metrics, so relying on a single metric is unsafe.

\subsubsection*{Comparing Explanation Method Families}

Beyond benchmarking individual metrics, practitioners must choose among families of explanation methods whose properties differ systematically. Table~\ref{tab:method_comparison} summarizes the qualitative characteristics of the method families encountered most frequently across the surveyed domains.

\begin{table*}[ht!]
\centering
\scriptsize
\renewcommand{\arraystretch}{1.1}
\caption{Qualitative Comparison of Explanation Method Families in XNLP}
\label{tab:method_comparison}
\begin{tabular}{p{2.6cm} p{1.9cm} p{4.6cm} p{2.2cm} p{2.9cm}}
\toprule
\textbf{Method Family} & \textbf{Scope} & \textbf{Faithfulness Evidence} & \textbf{Computational Cost} & \textbf{Typical Domains (Table~\ref{tab:cross_domain})} \\
\midrule

LIME~\citep{ribeiro_why_2016}
& Local
& Locally faithful surrogate by construction; sensitive to sampling choices and out-of-distribution perturbations~\citep{xu2024ffidelity, amparore2024comprehensive}
& High (perturbation sampling)~\citep{hassan2025model}
& Finance, social science, medicine \\
\addlinespace

SHAP~\citep{lundberg2017unified}
& Local, with global aggregation
& Game-theoretic consistency and additivity guarantees; empirical fidelity varies across metrics and modalities~\citep{zhang2023m4, amparore2024comprehensive}
& High (exact values exponential; approximations common)~\citep{hassan2025model}
& Finance, medicine, HR \\
\addlinespace

Attention visualization~\citep{vaswani2017attention}
& Local
& Contested: attention weights correlate with, but need not causally determine, predictions~\citep{jain_attention_2019}
& Low (byproduct of inference)
& Medicine, CRM, chatbots \\
\addlinespace

Gradient/saliency, e.g., Integrated Gradients~\citep{sundararajan2017axiomatic}
& Local
& Axiomatic attribution guarantees; rankings disagree with other attribution families across metrics~\citep{klein2024navigating}
& Low--medium (few forward/backward passes)
& Finance, social science \\
\addlinespace

Self-rationales / chain-of-thought~\citep{wei2022chain}
& Local
& Plausible but frequently unfaithful to the model's internal computation~\citep{turpin2023language, lanham2023measuring}
& Medium (additional generation)
& Chatbots, CRM, medicine \\

\bottomrule
\end{tabular}
\end{table*}

Across unified benchmarks (M$^4$~\citep{zhang2023m4}, LATEC~\citep{klein2024navigating}, BEExAI~\citep{sithakoul2024beexai}, and EvalxNLP~\citep{dhaini2025evalxnlp}), a consistent picture emerges: attribution methods disagree substantially across faithfulness metrics and modalities, attention weights are the least reliable family when read directly as explanations, and perturbation-based methods carry the highest computational cost. Method choice should therefore be reported together with the metric suite used to validate it.

\subsubsection*{Domain-Specific Evaluation Requirements}

Effective XNLP evaluation must account for domain-specific constraints and stakeholder needs. Table~\ref{tab:domain_eval} summarizes how evaluation requirements vary across application domains, complementing Table~\ref{tab:cross_domain}.

\begin{table*}[ht!]
\centering
\scriptsize
\renewcommand{\arraystretch}{1.1}
\caption{Domain-Specific Evaluation Requirements for XNLP}
\label{tab:domain_eval}
\begin{tabular}{p{1.8cm} p{2.8cm} p{2.5cm} p{3.5cm} p{2.5cm}}
\toprule
\textbf{Domain} & \textbf{Primary Metrics} & \textbf{Validation Method} & \textbf{Key Challenges} & \textbf{Stakeholders} \\
\midrule

\textbf{Medicine}
& AUC, clinical validity, fidelity, user trust
& Expert clinician review; clinical trial validation
& Regulatory approval (FDA/EMA); integration with clinical workflows; liability concerns~\citep{nazir2025xai}
& Clinicians, patients, regulators \\
\addlinespace

\textbf{Finance}
& Fidelity, compliance score, robustness
& Regulatory audit; stress testing; adversarial evaluation
& Adversarial robustness; temporal stability; real-time requirements~\citep{vcernevivciene2024explainable}
& Analysts, auditors, regulators \\
\addlinespace

\textbf{Systematic Reviews}
& WSS@95, recall, inter-rater agreement ($\kappa$)
& Benchmark replication; expert screening audit
& Heterogeneous corpora; transparent stopping criteria
& Reviewers, methodologists \\
\addlinespace

\textbf{Social \& Behavioral Science}
& Fairness, faithfulness, cultural validity
& Crowdsource annotation; expert review; cross-cultural validation
& Annotation bias; cultural variation in interpretation; platform specificity~\citep{gongane2025decoding}
& Researchers, moderators, users \\
\addlinespace

\textbf{HR}
& Fairness metrics, disparate impact, legal compliance
& Legal review; bias audits; demographic parity testing
& Legal liability; protected attribute handling; cross-jurisdictional requirements
& HR professionals, legal counsel, candidates \\
\addlinespace

\textbf{CRM}
& User satisfaction, trust, engagement
& A/B testing; user surveys; behavioral metrics
& Real-time generation; personalization trade-offs; multilingual support
& Customer service agents, end users \\
\addlinespace

\textbf{Chatbots}
& User trust, satisfaction, task completion
& A/B testing; user studies
& Real-time generation; conversational flow
& End users, service owners \\

\bottomrule
\end{tabular}
\end{table*}

These domain-specific requirements show that XNLP evaluation cannot rely solely on generic technical metrics. Healthcare applications require clinical validation that goes far beyond algorithmic fidelity, while financial applications must demonstrate robustness against adversarial manipulation. Future evaluation frameworks should incorporate domain-specific validation protocols alongside standardized technical metrics.

\subsubsection*{A Two-Tier Evaluation Protocol}

To make these requirements operational, we propose a two-tier protocol. Tier~1 comprises the core technical metrics that every XNLP study should report: fidelity (Eq.~\ref{eq:fidelity}), faithfulness assessed through perturbation or intervention tests, and completeness. Tier~2 is a domain validation layer chosen according to Table~\ref{tab:domain_eval}: expert clinical review in medicine, regulatory audit in finance, fairness audits in HR, and user studies in CRM and chatbot settings. In practice, a study would first identify the task and the stakeholders its explanations must serve. It would then report the Tier-1 technical metrics for the chosen explanation method, add the Tier-2 validation protocol appropriate to the deployment context, and disclose the evaluation artifacts (datasets, perturbation procedures, study instruments) so that others can replicate the assessment.

This protocol also answers a natural question raised by Table~\ref{tab:domain_eval}: domains do \emph{not} need wholly unique metric sets. They share a common technical core and differ principally in the validation layer through which that core is interpreted.

\subsection{Rationalization Techniques and Current Challenges}
\label{subsec:rationalization}

As NLP models become more intricate, providing explicit \emph{rationales} for their outputs has garnered growing attention~\cite{atanasova2024generating, rajani2019explain}. Often referred to as Rational NLP (RNLP), this subfield seeks to generate human-readable explanations of model decisions. While the foundational concept dates back to~\cite{zaidan2007using}, the surge in interest follows the wider availability of neural methods that can produce or highlight textual justifications.

\noindent
\textbf{Extractive Rationalization.} Techniques like LIME, Grad-CAM, or SHAP highlight the parts of the input most influential for a model's prediction~\cite{ribeiro_why_2016, lei_rationalizing_2016}, and information-theoretic variants select the smallest set of tokens sufficient to preserve the prediction~\cite{chen2018learning}. These methods are relatively straightforward to implement but can oversimplify the decision process. As models scale (e.g., large transformers), ensuring that these extractive saliency maps remain faithful is a growing challenge~\cite{majumder2021knowledge}.

\noindent
\textbf{Abstractive Rationalization.} Future directions in rationalization include generating free-form natural language explanations that may deviate from the exact phrasing of the input text~\cite{gurrapu_exclaim_2022, sha_rationalizing_2023}. This approach allows for more context-rich narratives but risks producing plausible yet inaccurate summaries if not rigorously tested for alignment with the model's internal states. Ribeiro et al.~\cite{ribeiro_semantically_2018} introduced ideas like Semantically Equivalent Adversarial Rules to refine rationalization by ensuring the model's textual explanations genuinely mirror its decision boundary.

Even so, post-hoc rationalizations can mislead if they do not reflect the true causal pathways in the model~\cite{turpin2023language}. Attribution-guided self-refinement offers one remedy, substantially reducing the unfaithfulness of LLM self-generated rationales~\citep{wang2025selfcritique}. Beyond faithfulness, widely used metrics often emphasize precision but rarely capture coherence, consistency, or domain relevance~\cite{jacovi2020towards}. Tackling these gaps is a precondition for dependable rationalization frameworks.

Two large-scale meta-evaluation suites are shaping how researchers assess XNLP methods. LATEC~\cite{klein2024navigating} benchmarks 17 explainers on 20 metrics across 7,500 settings, documents frequent disagreements between metrics, and argues for multi-faceted evaluation. BEExAI~\cite{sithakoul2024beexai} is an open-source library that computes a battery of quantitative metrics covering faithfulness, robustness, and complexity for a range of post-hoc explanation methods.
Complementary benchmarks continue to broaden this picture: FaithCoT-Bench provides an instance-level benchmark for detecting unfaithful chain-of-thought, comparing eleven detection methods across four domains~\citep{shen2025faithcot}, while SAEBench standardizes the evaluation of sparse autoencoders across eight metrics and hundreds of trained models~\citep{karvonen2025saebench}. On the user side, Kim et al.~\cite{kim2024human} reviewed 73 papers reporting 77 user studies of XAI, and argue that objective performance metrics have to be read alongside subjective user assessments. A companion review of 82 user studies with clinicians turns the same material into practical guidelines for human-centered evaluation~\citep{donosoguzman2025systematic}. Collectively, these efforts signal a shift toward reasoning-aware, user-centered evaluation protocols that extend beyond single numeric proxies.

\subsection{Data and Code Availability for Open Science}
\label{subsec:data_code_availability}

In parallel with technical innovations, open science principles shape the transparency and reproducibility of XNLP studies~\cite{brinkman2021open}. Public release of datasets and code makes rigorous validation possible, enabling other researchers to replicate, critique, or extend the work. Tools like Ecco~\cite{alammar-2021-ecco} help with interpreting transformer activations, aligning with open-source philosophies that lower barriers for scientific collaboration. Nonetheless, partial access or commercial constraints (e.g., proprietary GPT-4 models) can hamper in-depth analysis. During this survey, many XNLP projects provided open-source code on GitHub or Zenodo, but far fewer offered publicly accessible datasets, which remains a persistent bottleneck. S{\o}gaard~\cite{sogaard2021explainable} further highlights the value of standardized tasks and benchmarks, promoting fair comparisons across various XNLP domains. In that spirit, the tables in this survey, the XNLP taxonomy, and a curated index of explanation tools, benchmarks, and datasets are released as structured files in a companion repository\ifblind{} (anonymized for review: \url{\anonrepourl})\else{} at \url{\repourl}\fi{} so that they can be reused and extended.

\section{Future Directions and Research Opportunities in XNLP}
\phantomsection
\label{sec:future}

\noindent
\textbf{Reinforcement learning (RL) and Chain-of-Thought Reasoning.} RL increasingly intersects with advanced models like GPT, creating opportunities to refine both performance and interpretability~\cite{li_deep_2016}. Chain-of-thought prompting~\cite{wei2022chain}, wherein a model articulates its intermediate reasoning steps, can bolster interpretability, though these intermediate rationales are not always genuinely employed by the model's internal mechanics~\cite{lanham2023measuring} (see \emph{\nameref{subsec:cot_faithfulness}}). Future research may embed \emph{faithfulness tests} into the training objective, penalizing spurious or ungrounded reasoning steps; early work in this direction evaluates preference-optimization objectives such as GRPO and DPO for precisely that purpose~\citep{mohammadi2025faithfulcot}. A recent multi-laboratory position paper similarly frames CoT monitorability as a fragile explainability opportunity that training choices can inadvertently degrade~\citep{korbak2025chain}.

\noindent
\textbf{Hybrid Neuro-Symbolic Systems.} Combining neural networks with rule-based logic can produce interpretable, high-performing NLP pipelines. Weber et al.~\cite{weber_nlprolog_2019} illustrate a question-answering system that merges Prolog-style inferences with neural expansions, balancing explicit symbolic reasoning with the flexibility of learned representations. Similarly, Weir et al.~\cite{weir2024nellie} present NELLIE, which uses a Prolog-like proof tree guided by an LLM retriever, demonstrating how symbolic inference can offer auditable explanations while preserving neural adaptability.

\noindent
\textbf{Explainable Dialogue and Social Media Analytics.} Sarkar et al.~\cite{sarkar_towards_2023} review the explainability and safety of conversational agents built for mental health support, and identify where such systems still fall short of the transparency that setting demands. In social media contexts, XNLP can disclose the driving motivations behind viral content~\cite{bovet_influence_2019}, interpret emotional or political trends, and combat fake news. Here, personalized or context-driven explanations can be decisive in aligning NLP outputs with diverse user needs.

\noindent
\textbf{Personalized and Adaptive Explainability.} As user demographics and tasks grow more diverse, one-size-fits-all explanations may not suffice~\cite{kuhl_you_2020}. Systems like TELL-ME~\cite{jeck2025tell} let users specify whether they prefer analogical or factual styles, adapting future explanations accordingly. Tailoring the level of detail or communication style to match user domain expertise or cultural nuances can boost comprehension and adoption across varied environments.

\noindent
\textbf{Domain-Specific Opportunities.} The divergent requirements summarized in Table~\ref{tab:cross_domain} translate into concrete research agendas. In medicine, multimodal explanations over heterogeneous EHR data and drift-robust clinical validation protocols remain open problems. In finance, regulator-auditable explanation standards are needed as the EU AI Act and its accompanying General-Purpose AI Code of Practice come into force~\citep{dhaini2025from, europeancommission2025general}. In HR, emerging employment-AI regulation calls for bias-audit tooling that validates both fairness and competence of screening systems~\citep{webster2025fairness}. Social science applications need multilingual, culture-aware moderation explanations evaluated for both plausibility and faithfulness~\citep{rehman2026xmutest}. For systematic reviews, transparent stopping criteria and calibrated screening confidence are prerequisites for trustworthy end-to-end automation~\citep{cao2025automation}.

\noindent
\textbf{Rational AI (RAI).} Ongoing work in rational AI moves beyond surface-level justification, aiming to ensure that textual or visual rationales truly align with the model's underlying logic~\cite{yu_rethinking_2019}, directly addressing the faithfulness problem raised in \emph{\nameref{subsec:cot_faithfulness}}. Verified chain-of-thought reasoning, step-by-step evaluation, and semantically consistent rationales all converge to make model outputs not merely comprehensible but also trustworthy in real operational settings.

\noindent
In sum, XNLP's future hinges on bridging performance and transparency, using advanced techniques (e.g., chain-of-thought, RL) while upholding rigorous evaluation and open-science practices. Continued progress will likely depend on refining how we measure explanation quality, adapt them to user needs, and ensure that rationales faithfully mirror model processes.

\section{Conclusion}
\phantomsection
\label{sec:Conclusion}

In this paper, we examined XNLP by focusing on how it can be effectively applied across multiple domains to enhance user understanding, transparency, and trust in machine learning models. We began by examining the increasing sophistication of NLP systems and the intrinsic opacity of advanced architectures such as transformers. Particularly in high-stakes sectors like healthcare and finance, understanding \emph{why} a model predicts certain outcomes often matters as much as the predictions themselves.

We then traced the evolution of XNLP modeling techniques, starting from traditional methods like BoW and TF-IDF and advancing to embedding-based and transformer architectures. We highlighted various interpretability approaches, attention visualization, gradient-based explanations, and rationalization methods, that aim to demystify the inner workings of NLP systems. We also showed how these techniques can be used to align complex models with end-user requirements for clarity and reliability.

From here, we examined XNLP's implementation in distinct domains. In medicine, deployments in EHRs and clinical text analysis show how XNLP can generate actionable insights for patient care and build confidence among clinicians. In finance, XNLP supports risk assessment, fraud detection, and firm valuation, tasks in which accountability demands transparent predictive models. In systematic reviews, explaining which studies are included or excluded improves both the efficiency and the clarity of evidence-based research. In CRM, explainable sentiment analysis and automated summaries of customer feedback produce more trustworthy and user-friendly systems. Context-aware chatbots that justify their responses support user trust, provided the rationale is delivered without disrupting the dialogue. In social and behavioral science, XNLP helps detect hate speech, sexism, misinformation, and mental health signals while keeping the analysis of social data transparent and accountable. In HR, XNLP can automate tasks like recruitment and performance evaluation while offering explicit rationales that support fair and unbiased decisions. Beyond these domains, tasks like language generation, machine translation, and visual question answering demonstrate the breadth of XNLP's impact and the diverse challenges of making advanced models interpretable.

We then examined critical aspects of XNLP, discussing evaluation metrics (both quantitative and qualitative), potential trade-offs in model complexity and interpretability, and rationalization methods that strive for transparency while maintaining performance. We also highlighted the role of human-in-the-loop designs, where user feedback, bias detection, and user satisfaction can refine explanations, as well as the importance of open science practices for data and code availability. Finally, we outlined future research directions, including RL integrations, chain-of-thought reasoning, hybrid neuro-symbolic architectures, explainable dialogue systems, and personalized or adaptive explainability. These frontiers show how XNLP may further develop to meet real-world needs for trustworthy, user-aligned natural language solutions. Our cross-domain synthesis shows that trust is the convergent goal across all seven domains, whereas risk tolerance, temporal dynamics, stakeholder diversity, and explanation granularity diverge substantially. One-size-fits-all XNLP is therefore insufficient: what is needed is a shared technical core of explanation methods and metrics combined with a domain-specific validation layer.

\ifblind\else
  \section*{Acknowledgment}
The authors sincerely thank Tina Shahedi for her editorial assistance, helping refine the manuscript's language and clarity. We also appreciate the insightful feedback provided by Mehran Moazeni, Mohammad Behbahani, and Daniel Anadria, as well as the valuable suggestions from our anonymous reviewers.

\section*{Declarations}

\subsection*{Data availability}
No primary data were generated; all sources are cited in the reference list. The structured data tables, the XNLP taxonomy, and the tool and dataset index accompanying this survey are available at \url{\repourl}.

\subsection*{Conflict of interest}
The authors declare no potential conflicts of interest with respect to the research, authorship and/or publication of this article.

\subsection*{Funding sources}
This research did not receive any specific grant from funding agencies in the public, commercial, or not-for-profit sectors.

\fi

\appendix

\subsection{Glossary of Terms and Abbreviations}
\label{appendix:glossary}

Table~\ref{tab:glossary} lists the abbreviations and technical terms used throughout this survey.

\begin{table}[ht!]
\centering
\scriptsize
\renewcommand{\arraystretch}{1.1}
\caption{Common Definitions and Abbreviations}
\label{tab:glossary}
\begin{tabular}{p{2.5cm} p{4.5cm}}
\toprule
\textbf{Term/Abbrev.} & \textbf{Definition} \\
\midrule
\textbf{XNLP} & Explainable Natural Language Processing \\
\textbf{NLP} & Natural Language Processing \\
\textbf{AI} & Artificial Intelligence \\
\textbf{LLM} & Large Language Model \\
\textbf{CoT} & Chain-of-Thought (prompting/reasoning) \\
\textbf{SAE} & Sparse Autoencoder \\
\textbf{EHR} & Electronic Health Record \\
\textbf{CRM} & Customer Relationship Management \\
\textbf{HR} & Human Resources \\
\textbf{HITL} & Human-in-the-Loop (human feedback integrated into model training and validation) \\
\textbf{BoW} & Bag of Words \\
\textbf{TF-IDF} & Term Frequency--Inverse Document Frequency \\
\textbf{RNN} & Recurrent Neural Network \\
\textbf{RNLP} & Rational NLP (methods that generate explicit rationales for model decisions) \\
\textbf{BLEU} & Bilingual Evaluation Understudy (translation metric) \\
\textbf{GDPR} & General Data Protection Regulation \\
\textbf{HIPAA} & Health Insurance Portability and Accountability Act \\
\textbf{SVM} & Support Vector Machine \\
\textbf{CNN} & Convolutional Neural Network \\
\textbf{LSTM} & Long Short-Term Memory \\
\textbf{Transformer} & A neural network architecture using self-attention \\
\textbf{BERT} & Bidirectional Encoder Representations from Transformers \\
\textbf{DistilBERT} & A smaller, distilled version of BERT \\
\textbf{XLM-R} & A multilingual variant of RoBERTa \\
\textbf{BiGRU} & Bidirectional Gated Recurrent Unit \\
\textbf{BiLSTM} & Bidirectional Long Short-Term Memory \\
\textbf{Ensemble Model} & A combination of multiple models (e.g., BERT, XLM-R, DistilBERT) \\
\textbf{GPT-4} & Generative Pre-trained Transformer 4, a large language model by OpenAI \\
\textbf{MAE} & Mean Absolute Error \\
\textbf{AUC / AUC-ROC} & Area Under the ROC (Receiver Operating Characteristic) Curve \\
\textbf{Accuracy} & Ratio of correct predictions to total predictions \\
\textbf{Precision} & \(\frac{\text{True Positives}}{\text{True Positives} + \text{False Positives}}\) \\
\textbf{Recall} & \(\frac{\text{True Positives}}{\text{True Positives} + \text{False Negatives}}\) \\
\textbf{F1-score} & Harmonic mean of Precision and Recall \\
\textbf{RMSE} & Root Mean Squared Error \\
\textbf{ROUGE} & Recall-Oriented Understudy for Gisting Evaluation (a summarization metric) \\
\textbf{XAI} & Explainable Artificial Intelligence \\
\textbf{LIME} & Local Interpretable Model-agnostic Explanations \\
\textbf{SHAP} & SHapley Additive exPlanations \\
\textbf{MLQE-PE} & Machine Translation Quality Estimation - Post Editing dataset \\
\textbf{NMT} & Neural Machine Translation \\
\textbf{Probing Methodology} & Analyzing internal representations (e.g., attention) to assess model behavior \\
\textbf{RCT} & Randomized Controlled Trial \\
\textbf{SST-2} / \textbf{SST-5} & Stanford Sentiment Treebank (binary/5-class) \\
\textbf{UMNSRS} & University of Minnesota Semantic Relatedness Set \\
\textbf{WSS} & Work Saved over Sampling (systematic review metric) \\
\bottomrule
\end{tabular}
\end{table}
\subsection{Research Methodology}
\label{sec:appendixA}

\begin{figure*}[ht!]
\centering
\begin{subfigure}[b]{0.8\textwidth}
  \centering
  \includegraphics[width=\textwidth]{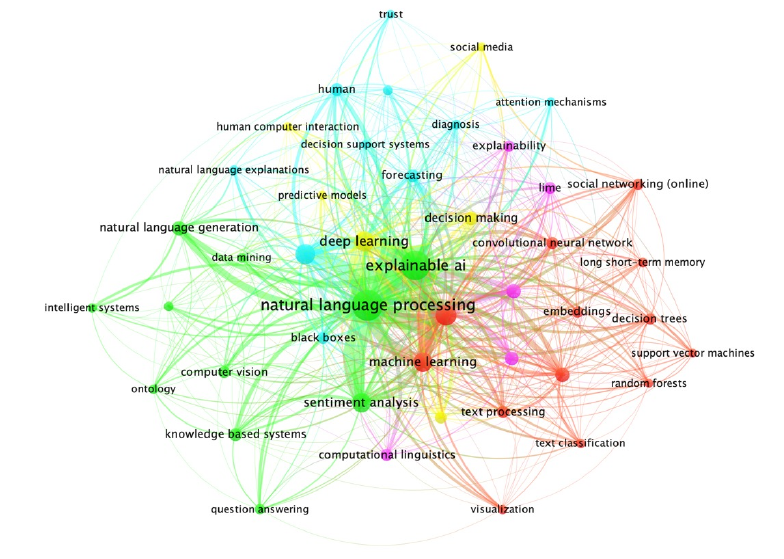}
  \caption{\small Keyword Network}
  \label{fig:VOSviewer}
\end{subfigure}

\vspace{1em}

\begin{subfigure}[b]{0.8\textwidth}
  \centering
  \includegraphics[width=\textwidth]{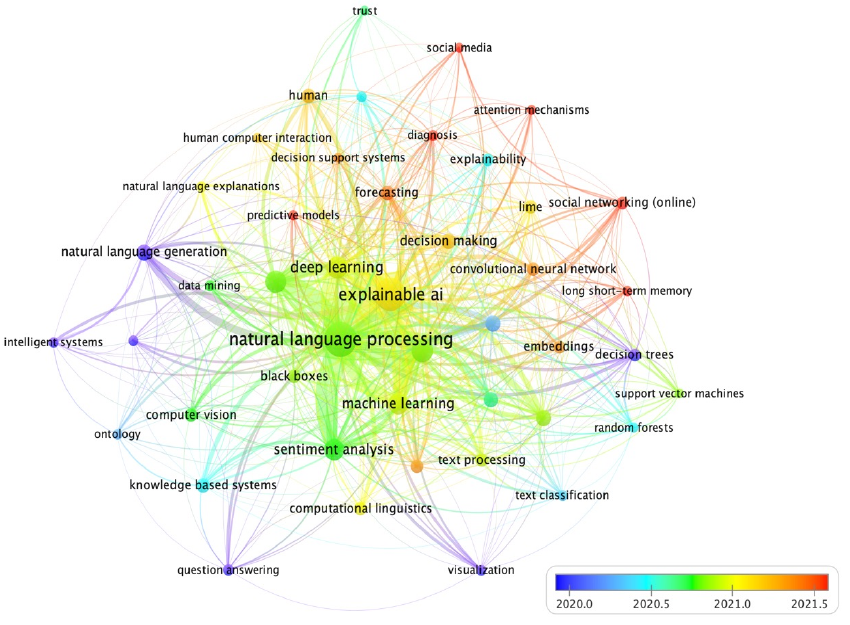}
  \caption{\small Overlay Visualization}
  \label{fig:VOSviewer2}
\end{subfigure}
\caption{\small VOSviewer views of the initial search corpus: (a)~network of co-occurring keywords; (b)~overlay map colored by average publication year (blue earlier, red more recent).}
\label{fig:VOSviewerFigures}
\end{figure*}

To initiate our review, we searched Google Scholar, IEEE Xplore, ACL, and ACM Digital Library for the terms ``explainable AI'' and ``natural language processing'' in titles, abstracts, and keywords. The search covered publications from 2018 onwards, and the reference list was later updated through 2026 during revision. This search provided a broad corpus of relevant academic work. We then refined the corpus through a bibliometric analysis with VOSviewer~\citep{vaneck2010software}, building keyword co-occurrence networks and identifying the major thematic clusters (Figure~\ref{fig:VOSviewerFigures}). This step produced a more coherent keyword set and yielded 217 candidate papers. After removing duplicates and marginally related items, 191 papers remained.

We then screened this cleaned dataset with ASReview, an open-source active-learning tool for systematic reviews~\citep{van_de_schoot_open_2021}, to find the articles most relevant to each XNLP application domain. This dual-stage approach, VOSviewer for bibliometric mapping and ASReview for targeted retrieval, proved effective in capturing the breadth and depth of XNLP research.

\noindent
Figure~\ref{fig:prisma} summarizes this selection pipeline in a PRISMA-style flow diagram. Cross-referencing the two tools sped up the literature analysis and clarified the focal points of XNLP research: the main ways XNLP is applied, the challenges observed, and the open questions that remain.

\begin{figure}[ht!]
\centering
\begin{tikzpicture}[
  node distance=3.5mm,
  pbox/.style={draw, rectangle, align=center, font=\scriptsize,
  text width=0.8\columnwidth, inner sep=4pt, fill=gray!5},
  phead/.style={font=\scriptsize\bfseries},
  parrow/.style={-stealth, thick}]
\node[pbox] (id) {\textbf{Identification}\\
  Keyword search (\emph{explainable AI}, \emph{natural language processing}) in titles, abstracts, and keywords across Google Scholar, IEEE Xplore, ACL, and ACM Digital Library (2018--2025)};
\node[pbox, below=of id] (bib) {\textbf{Bibliometric refinement (VOSviewer)}\\
  Keyword co-occurrence networks and thematic clusters\\
  $\rightarrow$ 217 candidate papers};
\node[pbox, below=of bib] (clean) {\textbf{Data cleaning}\\
  Duplicates and marginally related items removed\\
  $\rightarrow$ 191 papers};
\node[pbox, below=of clean] (asr) {\textbf{Active-learning screening (ASReview)}\\
  Iterative screening to isolate the most relevant articles per application domain};
\node[pbox, below=of asr, fill=gray!15] (incl) {\textbf{Included (191 papers; Table~\ref{tab:paper})}\\
  Medicine 26 $\cdot$ Finance 22 $\cdot$ Systematic Reviews 14 $\cdot$ CRM 31\\
  Chatbots 17 $\cdot$ Social/Behavioral Science 29 $\cdot$ HR 18 $\cdot$ Other 34};
\draw[parrow] (id) -- (bib);
\draw[parrow] (bib) -- (clean);
\draw[parrow] (clean) -- (asr);
\draw[parrow] (asr) -- (incl);
\end{tikzpicture}
\caption{\small PRISMA-style flow diagram of the literature selection pipeline: 217 candidate papers after bibliometric refinement, 191 after data cleaning, screened with ASReview into eight application domains (Table~\ref{tab:paper}).}
\label{fig:prisma}
\end{figure}

\noindent
Table~\ref{tab:paper} displays the final distribution of papers across different XNLP application domains, illustrating how we curated a representative yet succinct overview of current research. This approach streamlined the literature survey and let us focus more sharply on the domains covered in this review.

% \begin{table}[ht!]
% \centering
% \scriptsize
% \renewcommand{\arraystretch}{1}
% \caption{Number of Final Related Papers in Different Applications}
% \label{tab:paper}
% \begin{tabular}{p{1.2cm} p{5cm} p{1.2cm}}
% \toprule
% \textbf{No.} & \textbf{Application} & \textbf{No. Papers} \\
% \midrule
% 1 & Medicine & 12 \\
% 2 & Finance & 10 \\
% 3 & Systematic Reviews & 5 \\
% 4 & CRM & 12 \\
% 5 & Chatbots and Conversational AI & 5 \\
% 6 & Social and Behavioral Science & 11 \\
% 7 & HR & 7 \\
% 8 & Other Applications & 15 \\
% \bottomrule
% \end{tabular}
% \end{table}

\begin{table}[ht!]
\centering
\scriptsize
\renewcommand{\arraystretch}{1}
\caption{Number of Final Related Papers per Application Domain}
\label{tab:paper}
\begin{tabular}{p{1.2cm} p{5cm} p{1.2cm}}
\toprule
\textbf{No.} & \textbf{Application} & \textbf{No. Papers} \\
\midrule
1 & Medicine & 26 \\
2 & Finance & 22 \\
3 & Systematic Reviews & 14 \\
4 & CRM & 31 \\
5 & Chatbots and Conversational AI & 17 \\
6 & Social and Behavioral Science & 29 \\
7 & HR & 18 \\
8 & Other Applications & 34 \\
\bottomrule
\end{tabular}
\end{table}

% \subsection{Declaration of generative AI and AI-assisted technologies in the writing process}
% \label{sec:AI}

% During the preparation of this work, the authors used ChatGPT 3.5 from OpenAI to check grammar and make other writing corrections to improve the readability and language of the manuscript. After using this tool, the authors reviewed and edited the content as needed and took full responsibility for the published article.

\bibliographystyle{SageV}

\bibliography{base} % <-- This will use references.bib

% \begin{thebibliography}{99}
% \bibitem[Kopka and Daly(2003)]{R1}
% Kopka~H and Daly~PW (2003) \textit{A Guide to \LaTeX}, 4th~edn.
% Addison-Wesley.

% \bibitem[Lamport(1994)]{R2}
% Lamport~L (1994) \textit{\LaTeX: a Document Preparation System},
% 2nd~edn. Addison-Wesley.

% \bibitem[Mittelbach and Goossens(2004)]{R3}
% Mittelbach~F and Goossens~M (2004) \textit{The \LaTeX\ Companion},
% 2nd~edn. Addison-Wesley.

% \end{thebibliography}

\end{document}